\documentclass[10pt,twocolumn,letterpaper]{article}

\usepackage[pagenumbers]{cvpr} 

\usepackage[dvipsnames]{xcolor}
\usepackage{multirow}
\usepackage{makecell}
\usepackage{cuted}
\usepackage{tabularx}
\usepackage{chngcntr}
\usepackage{colortbl}
\usepackage{caption}
\usepackage{bm}
\usepackage{listings}
\usepackage{afterpage}

\usepackage[accsupp]{axessibility}  

\usepackage{tcolorbox}
\tcbuselibrary{listings, breakable}

\definecolor{lightgray}{gray}{0.95}

\newtcblisting{PromptBox}{
  colback=lightgray,
  colframe=black!30,
  listing only,
  listing options={
    basicstyle=\ttfamily\footnotesize,
    breaklines=true,
    columns=fullflexible,
    keepspaces=true,
    aboveskip=0pt,
    belowskip=0pt,
    lineskip=-1pt,
  },
  left=3pt,
  right=3pt,
  top=3pt,
  bottom=3pt,
  before=\setlength{\parskip}{0pt},
  before skip=6pt,
}

\newcommand{\firstcolor}{red!30}
\newcommand{\firstcell}[1]{\cellcolor{\firstcolor}#1}
\newcommand{\secondcolor}{orange!30}
\newcommand{\secondcell}[1]{\cellcolor{\secondcolor}#1}
\newcommand{\thirdcolor}{yellow!30}
\newcommand{\thirdcell}[1]{\cellcolor{\thirdcolor}#1}

\newcommand{\firstgray}{gray!50}
\newcommand{\firstgraycell}[1]{\cellcolor{\firstgray}#1}
\newcommand{\secondgray}{gray!30}
\newcommand{\secondgraycell}[1]{\cellcolor{\secondgray}#1}
\newcommand{\thirdgray}{gray!10}
\newcommand{\thirdgraycell}[1]{\cellcolor{\thirdgray}#1}

\definecolor{cvprblue}{rgb}{0.21,0.49,0.74}
\usepackage[pagebackref,breaklinks,colorlinks,allcolors=cvprblue]{hyperref}

\def\paperID{} 
\def\confName{CVPR}
\def\confYear{2026}

\title{

    Unified Camera Positional Encoding for Controlled Video Generation
}

\author{
    Cheng Zhang$^{1,2}$ \quad
    Boying Li$^{1}$\thanks{Corresponding author.} \quad
    Meng Wei$^{1}$
    \\
    Yan-Pei Cao$^{3}$ \quad
    Camilo Cruz Gambardella$^{1,2}$ \quad
    Dinh Phung$^{1}$ \quad
    Jianfei Cai$^{1}$
    \\
    $^{1}$Monash University \quad
    $^{2}$Building 4.0 CRC \quad
    $^{3}$VAST
}

\begin{document}
\maketitle
\begin{strip}
    \vspace{-3em}
    \centering
    \small
    \includegraphics[width=1.\linewidth, trim={0 0 0 0}, clip]{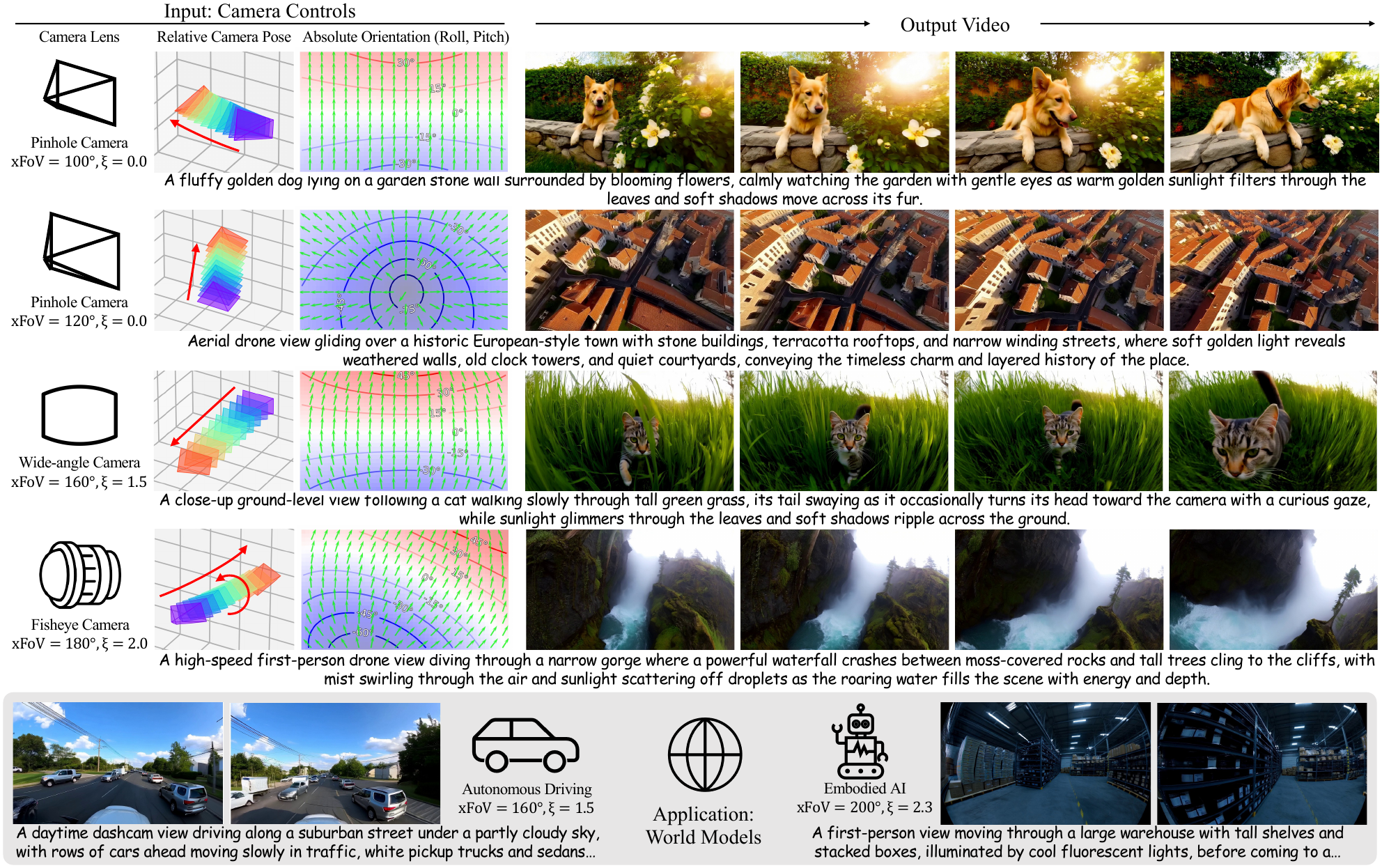}
    \vspace{-1.7em}
    \captionof{figure}{
        \textbf{Overview of our camera-controllable video generation.}
        Given user-specified text prompt and camera parameters including horizontal field-of-view, distortion \(\xi\), and camera poses with optional absolute orientation (encoded as latitude-up map), our model synthesizes realistic videos consistent with diverse camera geometries, demonstrating accurate pose and lens controllability and high visual fidelity.
        Applications span generative video content creation and world models for autonomous driving and embodied AI.
    }\label{fig:teaser}
    \vspace{-0.6em}
\end{strip}

\begin{abstract}
    Transformers have emerged as a universal backbone across 3D perception, video generation, and world models for autonomous driving and embodied AI, where understanding camera geometry is essential for grounding visual observations in three-dimensional space.
    However, existing camera encoding methods often rely on simplified pinhole assumptions, restricting generalization across the diverse intrinsics and lens distortions in real-world cameras.
    We introduce \textbf{Relative Ray Encoding}, a geometry-consistent representation that unifies complete camera information, including 6-DoF poses, intrinsics, and lens distortions.
    To evaluate its capability under diverse controllability demands, we adopt camera-controlled text-to-video generation as a testbed task.
    Within this setting, we further identify pitch and roll as two components effective for \textbf{Absolute Orientation Encoding}, enabling full control over the initial camera orientation.
    Together, these designs form \textbf{UCPE (Unified Camera Positional Encoding)}, which integrates into a pretrained video Diffusion Transformer through a lightweight spatial attention adapter, adding \textbf{less than 1\% trainable parameters} while achieving state-of-the-art camera controllability and visual fidelity.
    To facilitate systematic training and evaluation, we construct a large video dataset covering a wide range of camera motions and lens types.
    Extensive experiments validate the effectiveness of UCPE in camera-controllable video generation and highlight its potential as a general camera representation for Transformers across future multi-view, video, and 3D tasks.
    Code will be available at \url{https://github.com/chengzhag/UCPE}.
\end{abstract}
\vspace{-1em}
    
\section{Introduction}
\label{sec:intro}

Transformers~\cite{vaswani2017attention} have emerged as the foundation of modern architectures for novel view synthesis~\cite{jinlvsm}, 3D reconstruction~\cite{jiang2025rayzer}, and camera-controllable video generation~\cite{bahmani2025ac3d,he2024cameractrl,cheong2024boosting}, where networks must reason about how visual observations are formed by camera geometries (\eg, pose, intrinsics, projection model, lens distortion) in order to ground pixel sequences into 3D space.
To maximize visual coverage from limited viewpoints, real-world applications in autonomous driving~\cite{yogamani2019woodscape,kumar2020fisheyedistancenet}, robotic perception~\cite{mei2006calibration,nayar1997catadioptric,zhang2016benefit}, and world models~\cite{team2025hunyuanworld,bar2025navigation,yang2025matrix} routinely use fisheye, catadioptric, and equirectangular projections for 360° panoramas.
However, despite progress in Transformer backbones~\cite{jinlvsm,jiang2025rayzer,wan2025wan,yang2025cogvideox}, most camera-conditioned methods still rely on simplified pinhole assumptions, overlooking the highly nonlinear projections and strong lens distortions present in these diverse camera geometries.

\begin{figure}[t]
    \centering
    \includegraphics[width=1.\linewidth, trim={0 0 0 0}, clip]{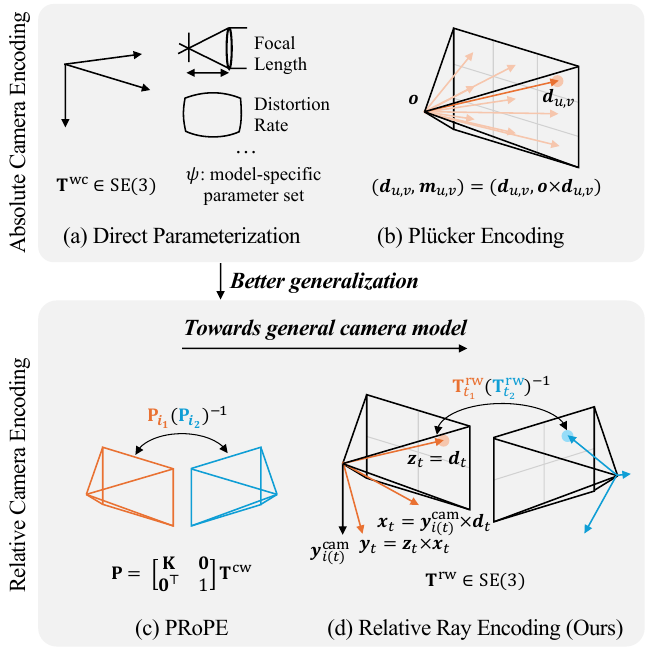}
    \vspace{-1.5em}
    \caption{
        \textbf{Comparison of camera encoding methods.}
        (a)~\emph{Direct Parameterization} encodes camera intrinsics and extrinsics as raw parameters, which lacks geometric interpretability and compatibility across camera types.
        (b)~\emph{Pl\"ucker Encoding} represents each ray as a pair of direction and moment vectors, providing a physically grounded but absolute, coordinate-dependent description.
        (c)~\emph{Projective Positional Encoding} encodes relative cameras in projective space, yet assumes pinhole projection and cannot model non-linear lens distortions.
        (d)~Our \emph{Relative Ray Encoding} reformulates geometric relationships in ray space, where each token corresponds to its own viewing ray, enabling better pose generalization and compatibility with \textbf{arbitrary} camera lenses.
    }\label{fig:rays}
    \vspace{-1em}
\end{figure}

In practice, these pinhole-based assumptions manifest in the camera representations fed into the network, as summarized in~\cref{fig:rays} and detailed in~\cref{sec:prelim}, which can be categorized into \emph{absolute} and \emph{relative} encodings.
Absolute encodings either feed raw camera parameters directly~\cite{bai2025recammaster,wang2024motionctrl} or adopt Pl\"ucker encoding~\cite{bahmani2025ac3d,he2024cameractrl} to represent rays under pinhole models.
However, as shown in our experiments (\cref{sec:cmp}), their dependence on specific world-frame configurations limits generalization across diverse camera geometries.
Recent relative encodings such as GTA~\cite{miyato2023gta} and PRoPE~\cite{li2025cameras} improve multi-view consistency in novel view synthesis by injecting geometry- or projection-aware relations into attention.
Yet, they remain restricted to the pinhole projection and are not readily compatible with pretrained video diffusion models.
\textit{Consequently, existing camera-conditioning methods still lack an encoding mechanism that explicitly models heterogeneous camera geometries, leading to limited generalization across varying intrinsics, distortion profiles, and viewpoint settings.}

To address these limitations, we propose \emph{Relative Ray Encoding} (\cref{fig:rays}d), which reformulates positional encoding from relative-camera encoding into relative-ray encoding.
By representing each token as a viewing ray expressed in local ray coordinates, this encoding enables the attention mechanism to operate directly in ray space rather than at the camera level, achieving geometry-consistent conditioning across heterogeneous camera lenses.

To evaluate this capability under diverse controllability demands, we adopt camera-controlled text-to-video generation as a testbed task.
However, existing camera-conditioned T2V methods~\cite{bahmani2025ac3d,he2024cameractrl} define camera poses only \emph{relative} to the first frame, leaving the global rotation frame ambiguous.
In particular, two degrees of freedom---\emph{pitch} and \emph{roll}---cannot be uniquely established, making it impossible to specify or reproduce the absolute orientation of the initial view.
To resolve this ambiguity, we introduce an \emph{Absolute Orientation Encoding} (third column of \cref{fig:teaser}), which anchors the camera to a gravity-aligned ``up'' direction.
This encoding explicitly provides absolute pitch and roll angles as a latitube-up map, enabling precise and repeatable control over the camera's absolute orientation.

Combining these two components, we arrive at \textbf{UCPE (Unified Camera Positional Encoding)}, a camera model-agnostic framework that injects complete camera geometry---including 6-DoF poses, intrinsics, and distortions---into Transformer attention.
UCPE provides a consistent formulation that unifies heterogeneous camera geometries, enabling fine-grained control over lens type, viewpoint, and orientation, as illustrated in~\cref{fig:teaser}.
Integrated into diffusion-based video generation through a lightweight spatial attention adapter, UCPE allows existing Diffusion Transformers to be fine-tuned with less than 1\% additional parameters.

To support systematic training and evaluation, we further construct a large-scale video dataset spanning diverse intrinsics, distortion profiles, and camera motions, offering a comprehensive benchmark for camera-controllable generation.
Experiments demonstrate that UCPE consistently improves camera controllability and visual fidelity across varied camera configurations, establishing a unified bridge between physical camera lenses and attention mechanisms for controllable video generation.

In summary, our contributions are as follows:
\begin{itemize}
    \item We propose \emph{UCPE}, a unified framework that encodes complete camera geometry (6-DoF poses, intrinsics, lens distortions) into Transformers. UCPE adopts a hybrid formulation combining \emph{Relative Ray Encoding} for ray-space geometric reasoning and projection nonlinearity and \emph{Absolute Orientation Encoding} for a gravity-aligned reference enabling controllable camera orientation.
    \item We introduce a lightweight \emph{attention injection method} that integrates UCPE into video Diffusion Transformers with less than 1\% of additional trainable parameters, enabling camera-aware video generation while preserving the visual fidelity of pretrained models.
    \item We construct a \emph{large-scale video dataset} covering diverse camera intrinsics, distortion profiles, and motion trajectories, providing a benchmark for evaluating camera-controllable generation under varied camera geometries.
    \item Extensive experiments demonstrate that UCPE not only enables accurate lens and orientation control, but also improves pose accuracy and generation quality for pinhole cameras and across diverse camera types.
\end{itemize}

\section{Related Work}
\label{sec:related}

\noindent\textbf{Video Diffusion Models.}
Diffusion-based generative models have become a powerful paradigm for video synthesis, achieving remarkable realism and temporal coherence.
Early attempts extended image diffusion frameworks~\cite{rombach2022high, guo2024animatediff} into the video domain by incorporating temporal modules to model motion dynamics.
Subsequent UNet-based~\cite{ho2022video, blattmann2023stable} and Transformer-based~\cite{yang2025cogvideox, wan2025wan} architectures operate in 3D video latent spaces to perform spatiotemporal denoising.
With the scaling up of training datasets and spatial resolution, large diffusion frameworks, such as HunyuanVideo~\cite{kong2024hunyuanvideo} and Seedance~\cite{gao2025seedance}, have pushed video generation quality to a level nearly indistinguishable from real-world footage.
To achieve controllability, these models typically incorporate textual, visual, or image-space conditions~\cite{wang2023videocomposer, guo2024sparsectrl, gu2025diffusion, burgert2025go} (\eg, depth, edges, or optical flow), to guide motion and appearance synthesis.
Nevertheless, most pretrained video diffusion frameworks remain agnostic to the underlying camera geometry that fundamentally determine how visual observations are formed.

\noindent\textbf{Camera-Controlled Generation.}
Recent efforts toward camera-controlled video generation aim to synthesize videos under explicit camera control, either through fine-tuning or training-free adaptation.
One line of work employs 3D representations to guide novel-view generation using image-space conditions such as rendering~\cite{yu2024viewcrafter,ma2025you,cao2025uni3c,zhang2025i2v3d,mengnvs,li2025realcam,zhai2025stargen,yu2025trajectorycrafter,i2vcontrolcamera}, tracking~\cite{gu2025diffusion}, optical flow~\cite{jin2025flovd,burgert2025go}, or coordinates~\cite{zhang2025world}.
These approaches facilitate viewpoint control and improve 3D consistency but depend on estimated depth maps, point clouds, meshes, or 3D Gaussians derived from input images or videos, thereby limiting motion diversity and often degrading under large camera movements or imperfect reconstructions.
Another direction conditions diffusion models directly on camera poses for video-to-video~\cite{bai2025recammaster}, image-to-video~\cite{liang2025wonderland,gao2024cat3d,zheng2024cami2v,zhou2025stable,he2025cameractrl}, and text-to-video generation~\cite{bahmani2025ac3d,he2024cameractrl,bahmanivd3d,wang2024motionctrl,cheong2024boosting}, relying on implicit 3D priors learned during training.
Although these methods demonstrate promising camera-aware generation, they typically define a reference frame on the first frame, without absolute orientation control (\eg, pitch and roll) in text-to-video synthesis.
Moreover, most existing approaches manipulate only extrinsic parameters (\ie, 6-DoF poses) while overlooking intrinsics and lens distortion, which hinders generalization across diverse camera configurations and projection models.
\textit{In contrast, our method introduces a unified, geometry-consistent representation that jointly encodes poses, intrinsics, and distortion, enabling physically grounded and accurate camera-aware control across a wide range of projection types.}

\noindent\textbf{Camera Encoding.}
Achieving fine-grained camera control in diffusion models fundamentally depends on how camera geometry is represented and encoded.
A few works explore direct camera conditioning by injecting camera parameters into diffusion models~\cite{wang2024motionctrl,bai2025recammaster}, enabling controllable viewpoint transitions without explicit 3D reconstruction.
Other methods encode camera poses through ray-map encodings for image-to-video~\cite{liang2025wonderland,gao2024cat3d,zheng2024cami2v,zhou2025stable,he2025cameractrl} and text-to-video generation~\cite{bahmani2025ac3d,he2024cameractrl,bahmanivd3d}.
Ray-map representations describe each pixel by its corresponding ray origin and direction~\cite{mildenhall2021nerf,gao2024cat3d} or by its Pl\"ucker coordinates~\cite{plucker1865xvii,bahmani2025ac3d,wang2025akira}, allowing models to incorporate both intrinsic and extrinsic camera properties.
However, these absolute encodings rely on a predefined world frame~\cite{mildenhall2019local,gao2024cat3d,guizilini2025zero}, making the representation dependent on arbitrary coordinate choices and limiting cross-scene generalization.

Motivated by the success of relative positional encodings such as RoPE~\cite{heo2024rotary}, several studies propose \textit{relative camera encodings} that model pairwise geometric transformations between views~\cite{safin2023repast,miyato2023gta,kong2024eschernet,li2025cameras}.
By encoding relative SE(3) transformations directly at the attention level, these approaches remove the need for a fixed reference frame and have been shown to improve multi-view reasoning and novel-view generation.
Building on these insights, \textit{our UCPE 
extends beyond perspective cameras by jointly encoding poses, intrinsics, and distortion within a geometry-consistent formulation.}
This unified representation seamlessly integrates with our attention adapter, allowing Diffusion Transformers to reason about camera geometry consistently across different projection types and achieve accurate viewpoint control.

\section{Method}
\label{sec:method}

\subsection{Preliminaries}
\label{sec:prelim}

\noindent\textbf{Camera as Ray Mapping.}
All camera models, regardless of their projection types, can be represented under a unified formulation that maps image coordinates to three-dimensional rays in space. 
Rather than assuming a specific projection model, a camera is defined by a ray-mapping function
\({\Phi}_{\psi}: ({u},{v}) \mapsto (\mathbf{o}_{{u},{v}}^{\textrm{cam}}, \mathbf{d}_{{u},{v}}^\textrm{cam})\),
which produces, for every pixel \(({u},{v})\), a ray origin \(\mathbf{o}_{{u},{v}}^{\textrm{cam}} \in \mathbb{R}^3\) and a unit direction \(\mathbf{d}_{{u},{v}}^{\textrm{cam}} \in \mathbb{S}^2\) in the camera coordinate system.
The mapping is parameterized by a model-specific set \(\psi\), \eg, focal length and distortion coefficients, depending on the chosen projection model.
For \emph{central} cameras, all rays share a common origin (\(\mathbf{o}_{{u},{v}}^{\textrm{cam}} = \mathbf{0}\)),
whereas \emph{non-central} cameras, such as catadioptric or panoramic systems, assign pixel-dependent origins. 
Unless otherwise stated, subsequent derivations assume a central camera model for notational simplicity.
We take the Unified Camera Model (UCM)~\cite{mei2007single} as a representative example (details in Supplementary~\cref{sec:syn_video}).

Let \(\mathbf{T}^{\textrm{wc}} = \left[\begin{smallmatrix} \mathbf{R} & \bm{t} \\ \bm{0}^\top & 1 \end{smallmatrix}\right] \in \mathrm{SE}(3)\) denote the camera pose from the camera to the world frame.
Applying this transformation yields the world-space ray representation:
\begin{equation}
\bm{d}_{{u},{v}} = \mathbf{R} \bm{d}_{{u},{v}}^{\textrm{cam}},
\qquad
\bm{o}_{{u},{v}} = \bm{t}.
\label{eq:ray_mapping}
\end{equation}
This 
provides a common geometric basis for reasoning about cameras with diverse projection characteristics.

\noindent\textbf{Absolute Camera Encodings}
{explicitly encode per-camera parameters or per-pixel rays in the world coordinate system.}
The most direct approach encodes raw camera parameters as numerical feature vectors~\cite{wang2024motionctrl,bai2025recammaster}, as shown in~\cref{fig:rays}a.
A more physically grounded alternative is the \emph{Pl\"ucker encoding}~\cite{bahmani2025ac3d,he2024cameractrl} (\cref{fig:rays}b), which reformulates each ray as a six-dimensional vector
\((\,\bm{d}_{{u},{v}},\, \bm{m}_{{u},{v}}\,) \in \mathbb{R}^6\),
where \(\bm{m}_{{u},{v}} = \bm{o}_{{u},{v}} \times \bm{d}_{{u},{v}}\) denotes the ray moment.
This representation provides a compact and numerically stable description of each pixel's ray, encoding both camera pose and lens in a physically interpretable form.
However, as an absolute encoding, it 
inherently remains sensitive to the chosen coordinate frame, which limits its generalizability across different scenes and viewpoints.

\noindent\textbf{Relative Camera Encodings.}
Recent work introduces relative encodings to model pairwise camera relationships directly within the attention mechanism of Transformers, making them invariant to global coordinate choices.
For each token \({t} \in \{1, \ldots, T\}\) with its corresponding camera index \({i}({t}) \in \{1, \ldots, N\}\),
a transformation matrix \(\mathbf{D}_{t}\) is derived from the world-to-camera transformation \(\mathbf{T}_{i({t})}^{\textrm{cw}} = (\mathbf{T}_{{i}({t})}^{\textrm{wc}})^{-1}\) as:
\begin{equation}
\mathbf{D}_{t} = \mathbf{I}_{{d}/4} \otimes \mathbf{T}_{{i}({t})}^{\textrm{cw}},
\label{eq:D_construct}
\end{equation}
where \(\mathbf{I}\) is the identity matrix, \({d}\) denotes the feature dimension, and \(\otimes\) is the Kronecker product that replicates the 3D transformation across feature subspaces.
The resulting block-diagonal matrix
\(\mathbf{D} = \operatorname{blkdiag}(\mathbf{D}_1,\ldots,\mathbf{D}_{T})\)
encodes per-token camera pose, which is applied to token features via token-wise matrix-vector multiplication \(\odot\),
\eg, \(\mathbf{D}\odot {Q} = \operatorname{blkdiag}(\mathbf{D}_1 {Q}_1, \ldots, \mathbf{D}_{T} {Q}_{T})\).

\emph{Camera Positional Encoding (CaPE)}~\cite{kong2024eschernet} employs such transformation matrices to inject relative camera poses into self-attention.
The attention operation is modified as:
\begin{equation}
{O}
= \operatorname{Attn}\big(\mathbf{D}^{\!\top}\!\odot {Q},\; \mathbf{D}^{-1}\!\odot {K},\; {V}\big).
\label{eq:cape}
\end{equation}
This replaces each query-key interaction \(Q_{{t}_1}^\top K_{{t}_2}\) with \({Q}_{{t}_1}^\top \mathbf{D}_{{t}_1} \! \mathbf{D}_{{t}_2}^{-1} {K}_{{t}_2}\),
where \(\mathbf{D}_{{t}_1} \! \mathbf{D}_{{t}_2}^{-1} = \mathbf{I}_{{d}/4} \!\otimes \mathbf{T}_{i({t}_1)}^{\textrm{cw}} \! (\mathbf{T}_{i({t}_2)}^{\textrm{cw}})^{\!-1}\),
conditioning attention on the relative camera pose.

\emph{Geometric Transform Attention (GTA)}~\cite{miyato2023gta} extends this formulation by additionally transforming the value matrix, thereby enforcing geometric consistency during both feature comparison and aggregation:
\begin{equation}
{O}
= \mathbf{D} \odot \operatorname{Attn}\big(\mathbf{D}^{\!\top}\!\odot {Q},\; \mathbf{D}^{-1}\!\odot {K},\; \mathbf{D}^{-1}\!\odot {V}\big).
\label{eq:gta}
\end{equation}

Building on this idea, \emph{Projective Positional Encoding (PRoPE)}~\cite{li2025cameras} (\cref{fig:rays}c) generalizes the relative transformation by replacing \(\mathbf{T}^{\textrm{cw}}_{i}\) with a camera projective matrix \(\mathbf{P}_{i} = \left[\begin{smallmatrix} \mathbf{K}_{i} & \bm{0} \\ \bm{0}^\top & 1 \end{smallmatrix}\right]\,\mathbf{T}^{\textrm{cw}}_{i}\) under intrinsics \(\mathbf{K}_{i}\).
This projective formulation captures the complete camera frustum geometry, but is limited to pinhole cameras.

\subsection{Unified Camera Positional Encoding}
\label{sec:ucpe}

\noindent\textbf{Relative Ray Encoding.}
Existing relative camera encodings assume that all image tokens within the same view share a single, linear projection function.
This assumption treats the entire image as one rigid entity, simplifying inter-camera geometric reasoning but ignoring intra-camera deviations in projection geometry.
Consequently, such camera-level encodings perform well under the idealized pinhole model, but struggle to generalize to real cameras exhibiting spatially varying projection behaviors such as lens distortions and wide field-of-view effects.

To accommodate non-linear projection models, we reformulate the encoding from a \emph{camera-to-camera} to a \emph{ray-to-ray} transformation, capturing fine-grained geometric variations across image tokens.
Specifically, for each image token $t$, we aim to construct a local ray-to-world matrix $\mathbf{T}^{\textrm{wr}}_t$, serving as the geometric operator for attention-level feature transformation.
Formally, each image token \(t\) is associated with a viewing ray parameterized by its origin and direction in the world coordinate system:
\begin{equation}
\bm{r}_t = \big(\bm{o}_t,\, \bm{d}_t\big),
\quad
\bm{o}_t \in \mathbb{R}^3,
\quad
\bm{d}_t \in \mathbb{S}^2,
\label{eq:ray_def}
\end{equation}
where \((\bm{o}_t, \bm{d}_t)\) is derived as in~\cref{eq:ray_mapping}.
We construct $\mathbf{T}^{\textrm{wr}}_t$ through defining a \emph{local ray coordinate system} anchored at the camera center, parameterized by an orthonormal basis \(\mathbf{R}^{\textrm{wr}}_t = [\mathbf{x}_t,\mathbf{y}_t,\mathbf{z}_t]\). As shown in~\cref{fig:rays}d, we take each ray direction $\mathbf{d}_t$ as the local $z$-axis and determine the other two orthogonal axes based on the camera's downward direction \(\mathbf{y}^{\textrm{cam}}_{i(t)}\) as follows:
\begin{equation}
\bm{z}_t = \bm{d}_t,
\quad
\bm{x}_t = \bm{y}^{\textrm{cam}}_{i(t)} \times \bm{z}_t,
\quad
\bm{y}_t = \bm{z}_t \times \bm{x}_t.
\label{eq:ray_axes}
\end{equation}
This orthonormal basis \(\mathbf{R}^{\textrm{wr}}_t\), together with the translation, \(\mathbf{t}^{\textrm{wr}}_t=\mathbf{o}_t\), forms a ray-to-world transformation:
\begin{equation}
\mathbf{T}^{\textrm{wr}}_t =
\begin{bmatrix}
\mathbf{R}^{\textrm{wr}}_t & \bm{t}^{\textrm{wr}}_t \\
\mathbf{0}^\top & 1
\end{bmatrix}.
\label{eq:ray2world}
\end{equation}
Each token feature is therefore associated with its specific geometric transform \(\mathbf{T}^{\textrm{rw}}_t = (\mathbf{T}^{\textrm{wr}}_t)^{-1}\),
mapping world coordinates into the local ray and serving as the geometric operator later for attention-level feature transformation. 
This formulation enables the attention mechanism to reason in ray space rather than sharing the same positional encoding within a camera frame, providing a consistent and physically interpretable geometric basis for UCPE both within and across frames.

\begin{figure}[t]
    \centering
    \includegraphics[width=0.9\linewidth, trim={0 0 0 0}, clip]{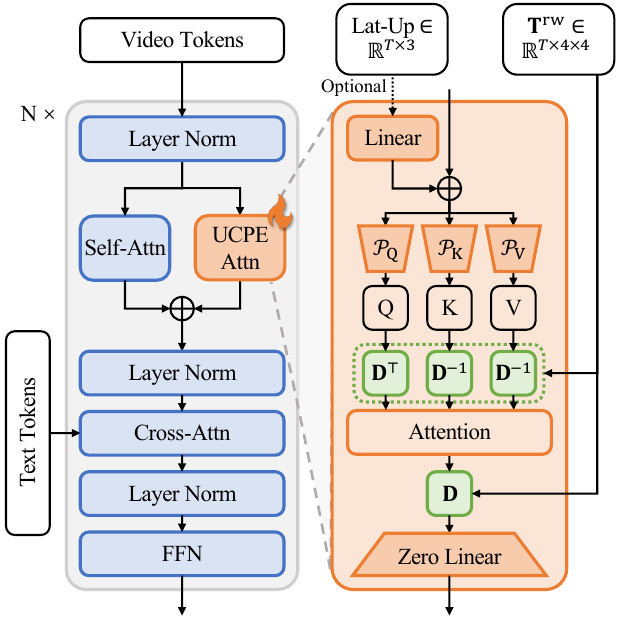}
    \vspace{-1em}
    \caption{
        \textbf{Overview of Spatial Attention Adapter.}
        The adapter injects UCPE into pretrained Transformers through a lightweight branch that preserves pretrained priors.
        It constructs hybrid encoding from the world-to-ray transform \(\mathbf{T}^{\textrm{rw}}\) and an optional Lat-Up map, applies them within attention, and fuses the resulting camera-aware tokens back through a zero-initialized linear layer.
    }\label{fig:pipeline}
    \vspace{-1em}
\end{figure}

\noindent\textbf{Absolute Orientation Encoding.}
While relative ray encoding enables geometry-consistent reasoning across heterogeneous cameras, it provides no notion of absolute orientation, leaving the first-frame orientation ambiguous.
Yet, most real-world videos are captured under a gravity-aligned ``up'' direction, which naturally defines absolute pitch and roll angles.
To incorporate this global orientation, we adopt the \emph{latitude-up map (Lat-Up map)}~\cite{jin2023perspective,veicht2024geocalib,bernal2025precisecam} shown in \cref{fig:teaser}, which concatenates each ray's \emph{latitude angle} and corresponding \emph{up-vector}.
This representation captures camera rotation through appearance cues such as sky-ground separation and object alignment, while also providing some distortion awareness for wide-angle lenses.

For the \emph{latitude map} calculation, we start from ray~\(t\)'s world-space direction \(\bm{d}_t = [d_{t,x}, d_{t,y}, d_{t,z}]^\top\) as defined in~\cref{eq:ray_def}.
The latitude map then encodes the angular elevation of each ray relative to the horizontal plane:
\begin{equation}
\text{Lat}_t = \arctan2 \big(-d_{t,y},\, \sqrt{d_{t,x}^{2} + d_{t,z}^{2}}\big),
\label{eq:latmap}
\end{equation}
where positive values correspond to upward-looking rays.

To compute the \emph{Up map}, we rotate each world-space ray \(\mathbf{d}_t\) by a small angle \(\delta\) around its local axis \(\bm{k}_t = \bm{d}_t \times \bm{u}^{\textrm{wld}}\),
where \(\bm{u}^{\textrm{wld}}\) denotes the world up direction.
The rotated ray \(\bm{d}^{\textrm{rot}}_t\) is then projected back to the image plane, and the resulting normalized pixel displacement defines the up direction:
\begin{equation}
\text{Up}_t =
\frac{[\Delta u_t,\; \Delta v_t]}{\|[\Delta u_t,\; \Delta v_t]\|},
\label{eq:upmap}
\end{equation}
where \([\Delta u_t,\Delta v_t]\) is the projection offset between the rotated and original rays (details in Supplementary~\cref{sec:abs_orient}).
The resulting Lat-Up encoding \([\text{Lat}_t, \text{Up}_t]\) provides each token with a global orientation context, enabling explicit control over camera pitch and roll during generation.

UCPE thus integrates the two complementary cues, \ie, local ray geometry and absolute orientation, to provide consistent and physically interpretable encoding that generalizes 
across camera lenses.

\subsection{Spatial Attention Adapter for UCPE Injection}
\label{sec:adapter}

\noindent\textbf{Spatial Attention with Hybrid Encoding.}
We follow~\cite{li2025cameras} to fuse the relative ray encoding with RoPE to promote both ray-space and image-space reasoning.
For each token \(t\), we construct a block-diagonal operator:
\begin{equation}
\mathbf{D}^{\textrm{UCPE}}_t
= \operatorname{blkdiag}\!\big(
\mathbf{D}^{\textrm{Ray}}_t,\;
\mathbf{D}^{\textrm{RoPE}}_t
\big),
\label{eq:ucpe_block}
\end{equation}
where \(\mathbf{D}^{\textrm{Ray}}_t = \mathbf{I}_{\frac{d}{8}}\!\otimes\!\mathbf{T}^{\textrm{rw}}_t \in \mathbb{R}^{\frac{d}{2}\times\frac{d}{2}}\)
encodes the world-to-ray transformation defined in~\cref{eq:ray2world},
and \(\mathbf{D}^{\textrm{RoPE}}_t \in \mathbb{R}^{\frac{d}{2}\times\frac{d}{2}}\)
is constructed with RoPE~\cite{heo2024rotary} to encode image-space positions, both with half of the feature dimension.
We share this design across GTA and PRoPE baselines, which is their best-performing configuration~\cite{li2025cameras}.
We then define \(\mathbf{D}^{\textrm{UCPE}}=\operatorname{blkdiag}(\mathbf{D}^{\textrm{UCPE}}_1,\ldots,\mathbf{D}^{\textrm{UCPE}}_T)\),
and apply it to the attention operation as shown in~\cref{fig:pipeline} following~\cref{eq:gta}.

Additionally, the Spatial Attention module takes a Lat-Up map as an optional input to provide absolute orientation control.
When present, the Lat-Up feature 
is projected by a linear layer to match the token dimension and then added to the input tokens as bias term.

\noindent\textbf{Adapting Transformer with Attention.}
We adopt Wan~\cite{wan2025wan} as our pretrained Diffusion Transformer, and intuitively the most suitable place to inject UCPE is within its self-attention layers that model spatial and temporal correlations.
However, directly replacing the existing positional encoding (\eg, 3D RoPE) with UCPE may disturb the priors established during large-scale pretraining. 

To preserve priors, we integrate UCPE through a lightweight, LoRA-inspired adapter that operates in parallel with the original attention. 
As in~\cref{fig:pipeline}, each DiT block retains standard self-attention while introducing a camera-conditioned branch \(\operatorname{UCPEAttn}(\cdot)\) based on the proposed encoding.
Crucially, thanks to the geometry-aware design of UCPE, this adapter requires only a small number of parameters to effectively model camera-conditioned correlations. 
In practice, we linearly project input tokens to \(1/C\) of the original dimension using \(\mathcal{P}_{Q}, \mathcal{P}_{K}, \mathcal{P}_{V}\), and proportionally reduce the number of attention heads, effectively reducing parameters and computation.
The \(\operatorname{UCPEAttn}(\cdot)\) output is then mapped back through a linear projection layer with zero-initialized weights, ensuring that the pretrained model remains unaltered at initialization.

The spatial adapter enables efficient integration of UCPE into Diffusion Transformers, providing fine-grained camera control with minimal additional parameters.

\section{Experiment}
\label{sec:experiment}

\begin{figure}[t]
    \centering
    \includegraphics[width=1.\linewidth, trim={0 0 0 0}, clip]{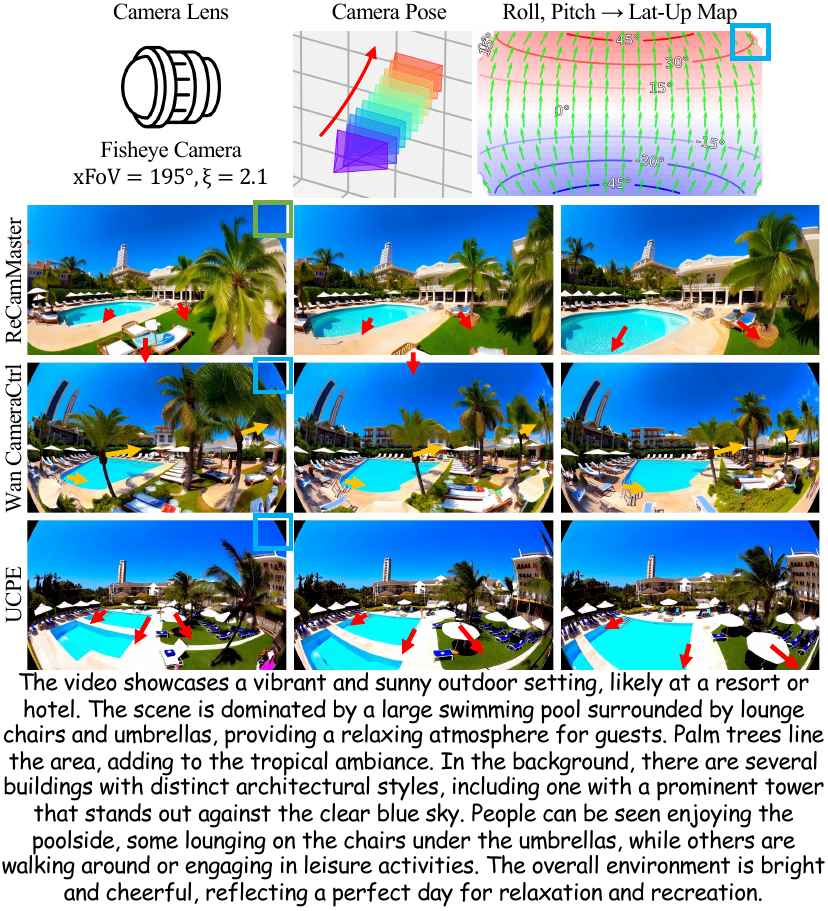}
    \vspace{-2em}
    \caption{
        \textbf{Comparison on our synthesized dataset.}
        UCPE \textcolor{red}{faithfully follows target trajectories} and produces \textcolor{cyan}{consistent lens distortions} aligned with visualization of the Lat-Up Map.
        In contrast, Wan CameraCtrl shows \textcolor{orange}{camera motion deviations}, while ReCamMaster \textcolor{ForestGreen}{fails to reproduce the intended distortion}.
        Colors correspond to the highlighted effects in the figure.
    }\label{fig:panshot}
    \vspace{-1em}
\end{figure}

\begin{figure*}[t]
    \vspace{-1em}
    \centering
    \includegraphics[width=1.\linewidth, trim={0 0 0 0}, clip]{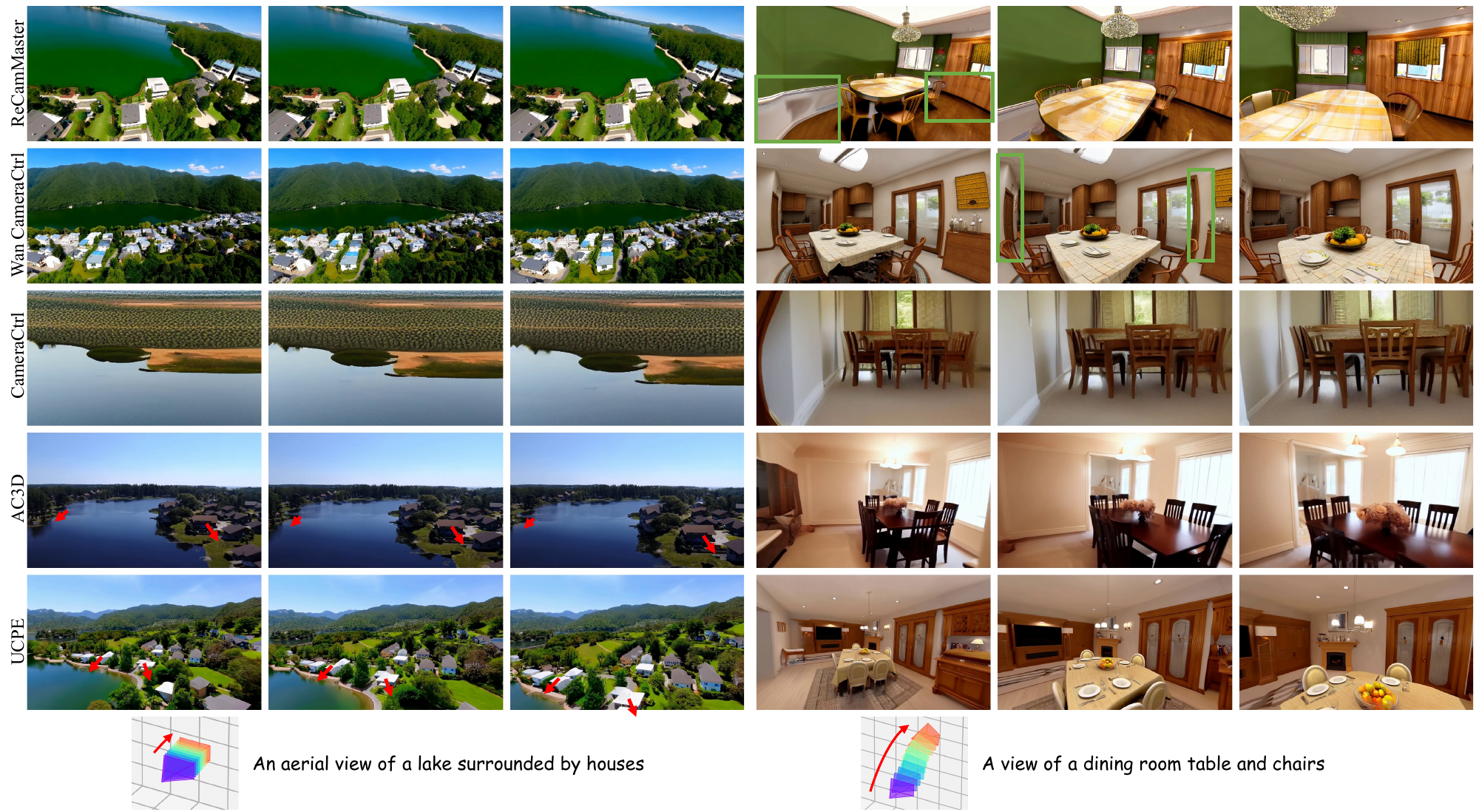}
    \vspace{-2em}
    \caption{
        \textbf{Comparison on the RealEstate10K dataset.}
        UCPE generates sharper, more detailed frames that better follow the target camera motion.  
        CameraCtrl produces severe artifacts (left) and poor composition (right), while AC3D preserves the training dataset's aesthetic but shows unbalanced framing (left) and low dynamic range (right).
        Wan CameraCtrl and ReCamMaster, though based the same backbone, struggle with camera consistency, leading to reduced motion (left) and \textcolor{ForestGreen}{undesired distortion artifacts} (right) under the pinhole setup.
    }\label{fig:re10k}
    \vspace{-1em}
\end{figure*}

\subsection{Experimental Setup}
\label{sec:setup}

\noindent\textbf{Datasets.}
To train and evaluate our method across diverse camera conditions (pose, intrinsics, and distortion) we synthesize a large-scale dataset of $\sim$48k clips from in-the-wild 360° videos~\cite{wallingford2024image} using Unified Camera Model (UCM)~\cite{mei2007single} with randomized $\mathrm{xFoV}$ and distortion $\xi$, covering pinhole, wide-angle, fisheye configurations (see Supplementary~\cref{sec:panshot}).
For out-of-distribution generalization, we additionally evaluate on 100 randomly selected clips from the RealEstate10K test set~\cite{zhou2018stereo}, using a fixed 100° \(\mathrm{xFoV}\) pinhole camera without any further adaptation or fine-tuning.

\noindent\textbf{Evaluation Metrics.}
We evaluate generated videos in four aspects (detailed in Supplementary~\cref{sec:eval}):
\begin{itemize}[leftmargin=0pt, wide=0pt]
    \item \emph{Video quality:} evaluated using FID~\cite{heusel2017gans}, FVD~\cite{unterthiner2018towards} for dataset-level fidelity, CLIP score~\cite{radford2021learning} for text-video alignment and Q-Align~\cite{wu2024q} for user-centric quality.
    \item \emph{Relative camera pose control:} each generated frame is rectified to a pinhole view with ground-truth distortion and estimated by ViPE~\cite{huang2025vipe}, reporting translation (TransErr), rotation (RotErr), and motion consistency (CamMC)~\cite{he2024cameractrl, wang2024motionctrl}.
    \item \emph{Absolute camera orientation:} pitch and roll are estimated via GeoCalib~\cite{veicht2024geocalib} and compared with the ground-truth video estimation to obtain absolute orientation errors.
    \item \emph{Lens control:} we calibrate FoV and distortion coefficients \(k_1, k_2\) of the radial model~\cite{Brown1966DecenteringDO} with GeoCalib~\cite{veicht2024geocalib} and compute absolute errors against ground-truth calibration.
\end{itemize}

\subsection{Comparison with Previous Methods}
\label{sec:cmp}

\noindent\textbf{Baselines.}
We compare UCPE against several representative camera-conditioning methods:
\begin{enumerate}[leftmargin=0pt, wide=0pt, label=(\arabic*)]
    \item \emph{ReCamMaster}~\cite{bai2025recammaster} injects camera extrinsics \(\left[\mathbf{R}, \mathbf{t}\right] \in \mathrm{SE}(3)\) into the diffusion model; we extend it by concatenating normalized \(\mathrm{FoV}\) and distortion \(\xi\) for lens control.
    \item \emph{Wan CameraCtrl} (Wan2.1-Fun-V1.1-1.3B-Control), a third-party implementation of CameraCtrl~\cite{he2024cameractrl} with Wan~\cite{wan2025wan}, which incorporates Pl\"ucker encoding via a convolutional adapter; we adapt it for text-to-video generation.
    \item \emph{AC3D}~\cite{bahmani2025ac3d} applies Pl\"ucker encoding through a ControlNet~\cite{zhang2023adding} adapter built on CogVideoX~\cite{yang2025cogvideox}.
    \item \emph{CameraCtrl}~\cite{he2024cameractrl} integrates Pl\"ucker encoding into U-Net's temporal attention layers of AnimateDiff~\cite{guo2024animatediff}.
\end{enumerate}
For both methods (1) and (2), we implement two variants respectively: one following their original design where camera poses are defined relative to the first frame, and another adopting a gravity-aligned coordinate system anchored at the first frame to enable explicit roll and pitch control (denoted as w/ absolute orientation).
Methods (1) and (2) share the same base model as UCPE and are fine-tuned on our dataset under identical settings for fair comparison, except that (2) uses a smaller learning rate of \(1\mathrm{e}{-5}\) due to full-parameter optimization (see Supplementary~\cref{sec:impl} for details).
For (3) and (4), we directly evaluate the pretrained models released by the authors, and therefore report results only on their training dataset, RealEstate10K~\cite{zhou2018stereo}.

\begin{table*}[t]
    \centering
    \vspace{-1em}
    \caption{
        \textbf{Quantitative comparison on our synthesized dataset.}
        UCPE outperforms all baselines across lens, orientation, and pose control while maintaining strong video quality with 90\% fewer parameters than ReCamMaster.
        Under both w/o and w/ Absolute Orientation Control, it yields lower pitch, roll, and rotation errors.
        Ablation results (bottom) show that moderate compression ratios in the Spatial Attention Adapter (e.g., 1/8-dim) best balance controllability and fidelity.
        Gray cells denote metrics not applicable due to missing control.
        Here, \emph{1/C-dim} denotes the token projection compression ratio, and \((d\times n)\) are the per-head dimension and number of attention heads.
    }\label{tbl:panshot}
    \vspace{-0.5em}

    \resizebox*{1.0\linewidth}{!}{
        \begin{tabular}{clcccccccccccc}
        \toprule
        \multirow[c]{2}{*}{\makecell[c]{Settings}} & \multirow[c]{2}{*}{Method} & \multicolumn{3}{c}{Camera Lens Control} & \multicolumn{2}{c}{Absolute Orientation} & \multicolumn{3}{c}{Relative Camera Pose Control} & \multicolumn{3}{c}{Video Generation Quality} & \multirow[c]{2}{*}{\makecell[c]{Trainable \\ Params}} \\
        \cmidrule(lr){3-5} \cmidrule(lr){6-7} \cmidrule(lr){8-10} \cmidrule(lr){11-13}
         & & FoV (°)$\downarrow$ & ${k}_{1}$$\downarrow$ & ${k}_{2}$$\downarrow$ & Pitch (°)$\downarrow$ & Roll (°)$\downarrow$ & RotErr (°)$\downarrow$ & TransErr$\downarrow$ & CamMC$\downarrow$ & FVD$\downarrow$ & FID$\downarrow$ & CLIP$\uparrow$ & \\
        \midrule
        \multirow[c]{3}{*}{\makecell[c]{w/o Absolute \\ Orientation \\ Control}} & ReCamMaster & \thirdcell 10.25 & \secondcell 0.210 & \thirdcell 0.143 & \thirdgraycell 10.00 & \thirdgraycell 7.42 & \secondcell 10.89 & \secondcell 31.44 & \secondcell 37.38 & \firstcell 555.54 & \thirdcell 69.91 & \thirdcell 24.86 & 
        354M \\
        & Wan CameraCtrl & \secondcell 10.05 & \thirdcell 0.222 & \secondcell 0.142 & \secondgraycell 9.35 & \secondgraycell 7.31 & \thirdcell 17.04 & \thirdcell 35.09 & \thirdcell 46.10 & \thirdcell 593.10 & \secondcell 67.83 & \secondcell 25.05 & 
        1.5B \\
        & UCPE & \firstcell 9.62 & \firstcell 0.174 & \firstcell 0.120 & \firstgraycell 8.03 & \firstgraycell 6.64 & \firstcell 4.29 & \firstcell 13.46 & \firstcell 15.94 & \secondcell 569.31 & \firstcell 66.22 & \firstcell 25.11 & 35.5M \\

        \midrule
        \multirow[c]{3}{*}{\makecell[c]{w/ Absolute \\ Orientation \\ Control}} & ReCamMaster & \thirdcell 10.04 & \secondcell 0.183 & \secondcell 0.136 & \thirdcell 6.62 & \secondcell 5.29 & \secondcell 9.23 & \secondcell 28.95 & \secondcell 33.88 & \thirdcell 605.83 & \thirdcell 67.07 & \thirdcell 24.84 & 
        354M \\
        & Wan CameraCtrl & \secondcell 9.86 & \thirdcell 0.230 & \thirdcell 0.162 & \secondcell 6.25 & \thirdcell 6.01 & \thirdcell 17.92 & \thirdcell 39.16 & \thirdcell 50.32 & \secondcell 554.43 & \secondcell 65.73 & \secondcell 25.01 & 
        1.5B \\
        & UCPE & \firstcell 8.22 & \firstcell 0.129 & \firstcell 0.102 & \firstcell 4.35 & \firstcell 3.74 & \firstcell 4.12 & \firstcell 15.21 & \firstcell 17.59 & \firstcell 495.14 & \firstcell 63.37 & \firstcell 25.12 & 35.6M \\

        \midrule
        \multirow[c]{8}{*}{\makecell[c]{w/ Absolute \\ Orientation \\ Control \\ \& \\ Ablation \\ Study}} & 1/2-dim ($128 \times 6$) & \secondcell 8.39 & 0.170 & 0.110 & 4.11 & 3.93 & \secondcell 3.69 & \firstcell 14.03 & \secondcell 16.06 & 534.44 & 64.88 & 25.05 & 141M \\
        & 1/4-dim ($128 \times 3$) & \thirdcell 8.47 & 0.149 & \firstcell 0.101 & \secondcell 3.94 & 4.08 & \firstcell 3.43 & \secondcell 14.26 & \firstcell 16.02 & 512.85 & \firstcell 62.86 & \secondcell 25.09 & 71.0M \\
        & 1/8-dim ($192 \times 1$) & \firstcell 8.22 & \firstcell 0.129 & \secondcell 0.102 & 4.35 & \thirdcell 3.74 & 4.12 & 15.21 & \thirdcell 17.59 & \secondcell 495.14 & \thirdcell 63.37 & \firstcell 25.12 & 35.6M \\
        & 1/12-dim ($128 \times 1$) & 8.96 & 0.151 & \thirdcell 0.107 & \firstcell 3.91 & 3.89 & 5.13 & \thirdcell 14.54 & 17.84 & \firstcell 487.54 & \secondcell 62.98 & 25.06 & 23.8M \\
        & Pre-Attn & \thirdcell 8.47 & \secondcell 0.145 & 0.108 & 4.26 & 3.96 & \thirdcell 4.03 & 15.77 & 17.85 & 502.73 & 63.62 & \thirdcell 25.07 & 35.6M \\
        & Post-Attn & 8.91 & \thirdcell 0.147 & 0.116 & \thirdcell 3.95 & 3.92 & 4.68 & 17.47 & 20.00 & 515.32 & 64.65 & 25.00 & 35.6M \\
        & PRoPE & 8.84 & 0.151 & 0.113 & 4.18 & \secondcell 3.70 & 5.35 & 17.52 & 20.58 & 516.59 & 65.00 & 25.03 & 35.6M \\
        & GTA & 8.80 & 0.157 & 0.117 & 4.21 & \firstcell 3.69 & 5.27 & 17.07 & 20.14 & \thirdcell 497.19 & 64.91 & 25.04 & 35.6M \\

        \bottomrule
        \end{tabular}
    }
    \vspace{-1em}
\end{table*}

\begin{table}[t]
    \centering
    \caption{
        \textbf{Quantitative comparison on RealEstate10K.}
        UCPE generalizes well without fine-tuning, achieving the lowest rotation, translation, and motion errors, and showing higher Q-Align scores than models trained on RealEstate10K (CameraCtrl and AC3D).
    }\label{tbl:re10k}
    \vspace{-0.5em}
        
    \resizebox{1.0\columnwidth}{!}{
        \begin{tabular}{lcccccc}
        \toprule
        \multirow[c]{2}{*}{Method} & \multicolumn{3}{c}{Relative Camera Pose Control} & \multicolumn{3}{c}{Q-Align Scores} \\

        \cmidrule(lr){2-4} \cmidrule(lr){5-7}

         & RotErr (°)$\downarrow$ & TransErr$\downarrow$ & CamMC$\downarrow$ & \makecell[c]{Image \\ Quality}$\uparrow$ & \makecell[c]{Image \\ Aesthetic}$\uparrow$ & \makecell[c]{Video \\ Quality}$\uparrow$ \\
        \midrule
        ReCamMaster & \thirdcell 1.10 & 5.64 & 6.15 & \secondcell 0.9492 & \secondcell 0.5185 & \secondcell 0.9720 \\
        Wan CameraCtrl & 2.22 & 7.42 & 8.67 & \firstcell 0.9822 & \firstcell 0.5691 & \firstcell 0.9885 \\
        CameraCtrl & 1.17 & \thirdcell 3.96 & \thirdcell 4.59 & 0.6877 & 0.3306 & 0.7338 \\
        AC3D & \secondcell 0.62 & \secondcell 2.11 & \secondcell 2.43 & 0.7699 & 0.3651 & 0.8211 \\
        UCPE & \firstcell 0.56 & \firstcell 1.25 & \firstcell 1.58 & \thirdcell 0.9480 & \thirdcell 0.4686 & \thirdcell 0.9694 \\

        \bottomrule
        \end{tabular}
    }
    \vspace{-1em}
\end{table}

\noindent\textbf{Quantitative Results.}
We present quantitative comparisons on our synthesized dataset in Table~\ref{tbl:panshot} and on RealEstate10K in Table~\ref{tbl:re10k}.
On our dataset, we first compare with ReCamMaster and Wan CameraCtrl under their original settings (w/o Absolute Orientation Control), and then under our proposed gravity-aligned coordinate system (w/ Absolute Orientation Control).
UCPE consistently achieves better performance across both configurations, exhibiting superior camera controllability and video quality while adding only 35.5M or 0.5\% parameters over the 7.3B parameters of the base model, 90\% fewer than ReCamMaster.
The Lat-Up map also provides appearance cues that improve lens control (UCPE w/ vs. w/o Absolute Orientation).

On RealEstate10K, despite not being fine-tuned on this dataset, UCPE outperforms all baselines in Relative Camera Pose Control, demonstrating strong generalization to unseen trajectories and text prompts.
Note that UCPE is trained with detailed scene descriptions (see \cref{fig:panshot}) whereas RealEstate10K uses short prompts (see \cref{fig:re10k}).
Nevertheless, UCPE achieves higher Q-Align scores than CameraCtrl and AC3D, indicating better video quality, and comparable performance to ReCamMaster and Wan CameraCtrl with fewer parameters.

\noindent\textbf{Qualitative Results.}
We visualize generated samples from our synthesized dataset in \cref{fig:panshot} and from RealEstate10K in \cref{fig:re10k}.
On our dataset, UCPE \textcolor{red}{faithfully follows the specified camera trajectories} and produces \textcolor{cyan}{consistent lens distortion effects} aligned with the target camera parameters visualized in the Lat-Up Map.
In contrast, Wan CameraCtrl exhibits noticeable \textcolor{orange}{camera motion deviations}, while ReCamMaster \textcolor{ForestGreen}{fails to reproduce the intended lens distortion}.
On RealEstate10K, UCPE generates sharper and more detailed videos that better adhere to the camera motion, whereas CameraCtrl shows severe artifacts (left) and poor frame composition (right).
AC3D maintains the overall aesthetic of RealEstate10K due to training on this dataset but suffers from unbalanced framing (left) and low dynamic range (overexposed window, right).
Although Wan CameraCtrl and ReCamMaster are fine-tuned on the same base model and dataset as ours, they struggle to ensure camera consistency, resulting in reduced camera motion (left) and \textcolor{ForestGreen}{undesired distortion artifacts} (right) under the pinhole setup.

\subsection{Ablation Study}

We conduct ablation experiments on our synthesized dataset to analyze the core designs 
of UCPE and the Spatial Attention Adapter, with results summarized in 
Table~\ref{tbl:panshot} (bottom).

\noindent\textbf{Compression Ratio of Token Projection.}
As introduced in~\cref{sec:adapter}, we apply a token projection with compression ratio \(1/C\) in the Spatial Attention Adapter to balance efficiency and representation capacity.
We test four ratios, each corresponding to different per-head dimensions and numbers of attention heads.
Higher compression ratios improve video quality metrics while maintaining stable camera controllability.
We adopt \(1/8\) as the default setting, which provides the best trade-off between performance and efficiency.

\noindent\textbf{Injection Position of the Attention Adapter.}
We compare three configurations within the Diffusion Transformer: insertion before self-attention (Pre-Attn), after self-attention (Post-Attn), and our default parallel design (1/8-dim).
Both Pre-Attn and Post-Attn variants yield noticeably degraded results in both camera control and video quality, confirming the effectiveness of our parallel architecture.

\noindent\textbf{Comparison with Other Relative Camera Encodings.}
To assess the contribution of our relative ray encoding, we replace it with prior formulations including PRoPE~\cite{li2025cameras} and GTA~\cite{miyato2023gta}, while retaining the Lat-Up Map for orientation and lens cues.
Compared to UCPE (1/8-dim), both alternatives lead to weaker camera lens and relative pose control, as well as lower video quality, though they achieve similar absolute orientation performance via the Lat-Up Map.
We attribute this to their limited capacity to model non-linear lens projections: both methods use a single camera encoding for all tokens, lacking the ability to capture distortion variation.
This limitation is especially pronounced for PRoPE, which even underperforms GTA on several metrics, highlighting the advantage of UCPE's relative ray encoding as a unified representation across diverse camera lens.

\section{Conclusion}
\label{sec:conclusion}

We presented UCPE, which jointly models poses, intrinsics, and lens distortions through relative ray encoding and absolute orientation cues.
Integrated via a lightweight spatial attention adapter, it enables video Diffusion Transformers to achieve accurate camera control with under 1\% additional parameters.
Experiments on our camera-diverse dataset show clear gains in lens controllability, orientation accuracy, and pose fidelity.
Overall, UCPE provides a unified representation for diverse cameras, potentially as a general encoding across multi-view, video and 3D tasks.

\noindent\textbf{Acknowledgement:}
This research is supported by Building 4.0 CRC and partially supported by the Australian Research Council under Grant DP260100218.

{
    \small
    \bibliographystyle{ieeenat_fullname}
    \bibliography{main}
}

\clearpage
\setcounter{page}{1}
\maketitlesupplementary

\counterwithin{figure}{section}
\counterwithin{table}{section}
\counterwithin{equation}{section}
\renewcommand\thesection{\Alph{section}}
\renewcommand\thetable{\thesection.\arabic{table}}
\renewcommand\thefigure{\thesection.\arabic{figure}}
\renewcommand\theequation{\thesection.\arabic{equation}}
\setcounter{section}{0}

\renewcommand{\theHsection}{supp.\Alph{section}}
\renewcommand{\theHtable}{supp.\Alph{section}.\arabic{table}}
\renewcommand{\theHfigure}{supp.\Alph{section}.\arabic{figure}}
\renewcommand{\theHequation}{supp.\Alph{section}.\arabic{equation}}

This supplementary material is organized as follows:
\begin{itemize}
    \item \cref{sec:panshot} provides additional details on our construction of the camera-diverse dataset.
    \item \cref{sec:abs_orient} elaborates the derivation of the absolute orientation encoding.
    \item \cref{sec:eval} summarizes the evaluation metrics used in our experiments.
    \item \cref{sec:impl} presents implementation details for both our pipeline and the baselines.
    \item \cref{sec:limitation} discusses the robustness, generalization capability, limitations, and future work.
    \item \cref{sec:demo} includes a detailed demo video.
\end{itemize}

\section{Dataset Construction}
\label{sec:panshot}

The main paper introduced UCPE, a unified camera encoding that jointly models 6-DoF poses, intrinsics, and lens distortions.  
Training and evaluating such a representation in a camera-controllable video generation setting requires large-scale data with diverse motions, FoVs, and distortions.  
However, collecting such real-world data is extremely expensive due to the need for multiple lenses, hardware configurations, and extensive manual effort.

To overcome these limitations, we synthesize camera-diverse videos by using 360° panoramic footage as a flexible exploration space, from which videos can be projected into arbitrary virtual cameras.

We design a \emph{three-stage data synthesis pipeline} comprising:
(1) \emph{360° panoramic video curation} to extract high-quality, gravity-aligned exploration clips;
(2) \emph{realistic rotation emulation} by transferring camera motions from perspective videos with similar translations; and
(3) \emph{camera-diverse video synthesis} under varied intrinsics, distortions, and augmented rotations.  
While Sec.~4.1 of the main paper provides an overview, here we present the full details.

\subsection{Coordinate Conventions}
\label{sec:coord}

Throughout this supplementary material, we adopt a consistent 3D coordinate convention for all projection, ray-mapping, and pose-related computations.
The coordinate frame is defined as follows:
\begin{itemize}
    \item the \(x\)-axis points to the right of the image;
    \item the \(y\)-axis points downward (following image coordinates);
    \item the \(z\)-axis points forward, aligned with the viewing direction of the ERP camera.
\end{itemize}

Under this convention, the world-up direction is
\begin{equation}
\bm{u}^{\mathrm{wld}} = [0,\,-1,\,0]^\top,
\label{eq:world_up}
\end{equation}
which is used consistently across ERP projections, UCM ray mappings
(Sec.~\ref{sec:panshot}), and the construction of the Up map
(Sec.~\ref{sec:abs_orient}).

\subsection{360° Panoramic Video Curation}
\label{sec:panflow}

A key benefit of UCPE is controllability over absolute pitch and yaw, resolving the ambiguity in conventional camera-controlled T2V generation.  
This requires all training videos to share a gravity-aware world frame, which is difficult to obtain from in-the-wild footage.

Fortunately, many commercial 360° cameras produce stabilized, gravity-aligned equirectangular panoramas using calibrated lenses and onboard IMUs.
This ensures a consistent up vector, enabling virtual cameras defined inside the panoramic sphere to inherit absolute pitch and yaw.

We begin from the large-scale 4K resolution 360° video corpus of~\cite{wallingford2024image}, containing 24.1k YouTube videos with diverse scenes and motions.  
These raw videos contain issues such as scene cuts, watermarks, low-quality footage, non-equirectangular projections, or imperfect stabilization.
We therefore construct a multi-stage pipeline to filter low-quality clips and obtain high-quality gravity-aligned panoramas suitable for camera-diverse video synthesis.

\paragraph{Clip Segmentation.}
Video generation models require temporally consistent clips.  
We first apply a threshold-based transition detector~\cite{scenedetect2024} to remove hard cuts and black fades, but this method often misses subtle transitions (\eg, cross fades).  
We then run panoramic visual SLAM~\cite{sumikura2019openvslam} on each preliminary segment.  
Empirically, SLAM fails to track features across scene transitions, leading to ``tracking-lost'' signals.
By monitoring these signals, we split the preliminary clips into more refined segments with consistent scenes, which results in roughly 400k clips, each of 10 seconds.

\paragraph{Camera Pose and Rotation Score Extraction.}
We run second pass of SLAM in localization mode to obtain more accurate poses, yielding approximately 300k clips with valid estimates.
As discussed earlier, enabling absolute orientation supervision requires removing clips that are unstabilized or not gravity-aligned.  
Since stabilized 360° videos are typically gravity-consistent, we examine the rotational components of the estimated poses and discard clips exhibiting large drift.
To quantify this, we compute a rotation score defined as the maximum relative rotation between each frame and the first frame:
\begin{equation}
    \alpha_{\max}
    =
    \max_{i>0}
    \arccos\!\left(
        \frac{\mathrm{Tr}\!\left(\mathbf{R}_0^\top \mathbf{R}_i\right)-1}{2}
    \right),
\end{equation}
where $\mathbf{R}_0$ is the first-frame rotation and $\mathbf{R}_i$ denotes the rotation at frame $i$.
Clips with excessive drift are later removed in~\cref{sec:emulate_rot}, ensuring consistent gravity-aware orientations.

\paragraph{Quality Filtering.}
The raw 360° videos contain various artifacts, which we address using three complementary filtering strategies:
\begin{itemize}
    \item \textbf{Low-quality filtering:} We use Q-Align~\cite{wu2024q} to assess image quality, aesthetics, and video quality, and average these scores as a quality indicator for candidate selection in~\cref{sec:emulate_rot}.
    \item \textbf{Watermark filtering:} A watermark detector~\cite{watermark} is applied to each frame, and the averaged score is used as a watermark indicator for candidate selection.
    \item \textbf{vLLM-based filtering:} A custom vLLM-based filter~\cite{bai2025qwen2} (see~\cref{fig:filter}) identifies and removes clips that are non-equirectangular, low-quality, or contain overlays, missing edges, or other artifacts.
    The vLLM model also assigns each clip to most relevant POI (Point of Interest) categories~\cite{xia2025panowan} for later semantic balancing.
\end{itemize}

\paragraph{Trajectory Scale Normalization.}
Monocular SLAM recovers camera trajectories only up to an arbitrary scale, causing the same unit translation to appear faster in shallow scenes and slower in deeper ones.  
To ensure consistent motion across clips, we estimate a per-clip geometric scale and normalize all trajectories accordingly.
Specifically, we compute dense optical flow using PanoFlow~\cite{shi2023panoflow}, recover per-pixel depth via generalized epipolar geometry, and obtain a near-plane statistic by taking the 25th percentile of valid depths per frame and then the median across the clip.
We rescale each trajectory by this value so that the median near-plane depth becomes~1, yielding consistent apparent motion across all training clips.

\subsection{Emulating Realistic Camera Rotations}
\label{sec:emulate_rot}

The curated panoramic clips serve as exploration spaces with provided camera translations, but determining realistic camera rotations remains challenging.
Synthetic rotations (e.g., constant-speed sweeps) appear unnatural and do not reflect real panning, tilting, or rolling patterns that correlate with translation.

Our key insight is that videos with similar translation motions tend to exhibit similar rotation dynamics.
Thus, we employ a matching-based approach with two steps:
(1) extraction of realistic rotation trajectories from perspective videos, and
(2) matching and transferring these rotations to panoramic clips based on translation similarity and clip quality.

\paragraph{Candidate Rotation Extraction.}
We use CameraBench~\cite{lin2025towards}, a dataset of 1k motion-diverse cinematic videos with text-described camera motions.  
We remove clips with ``zoom'' in desciptions to maintain fixed intrinsics, and extract camera poses using ViPE~\cite{huang2025vipe}, producing around 300 unique trajectories.  
In some challenging scenarios, ViPE may produce unstable or jittery poses.
To remove these results, we compute each clip's maximum rotational velocity and discard the top 20\%.  
We then align all trajectories to a gravity-aware world frame via GeoCalib~\cite{veicht2024geocalib} by aligning the first-frame up vector to the world up axis.
This step ensures that all candidate rotations can be directly applied to the gravity-aligned panoramic clips to produce natural absolute orientations.

\paragraph{Trajectory Matching and Rotation Transfer.}
For each panoramic clip, we find CameraBench trajectories with similar translations and transfer their rotations to the virtual camera.
We first align each candidate trajectory to the panoramic clip via Umeyama fitting~\cite{umeyama2002least}, constrained to vertical-axis rotation.
Then we retain the top 30 candidates with lowest RMSE error for further selection.

The original 360° videos exhibit serious long tails in POI categories (see~\cref{fig:category_before}), possibly due to the widespread use of 360° cameras for specific scenarios such as outdoor sports in the mountains and street views.
Therefore we apply a series of diversity constraints during candidate selection to ensure semantic balance and motion diversity.
Specifically, each candidate is scored using a composite metric averaging quality score, watermark score, and the rotation score $\alpha_{\max}$ from~\cref{sec:panflow}.
Then they are traversed greedily from best to worst while enforcing the following diversity constraints:
\begin{itemize}
    \item \textbf{Panoramic clip diversity:} At most 5 CameraBench matches per panorama.
    \item \textbf{CameraBench trajectory diversity:} Each trajectory can be used at most 100 times.
    \item \textbf{POI semantic balance:} Matches are skipped if all POI categories of the panoramic clip already exceed 1000 matches.
    \item \textbf{Motion diversity:} We compute the average optical-flow magnitude as a motion score and maintain bins of fixed width of 10, each capped at 2000 samples.
\end{itemize}

While these constraints do not guarantee perfect uniformity, they substantially reduce long-tail distributions in POI categories (see~\cref{fig:category_after}) and motion.
After selection, each panoramic clip is paired with 5 per-frame camera rotations aligned to equirectangular coordinate system, yielding a total of 12k pairs of panoramic clips and rotation sequences.

\afterpage{
    \clearpage
    \begin{figure*}[t]
\centering
\begin{PromptBox}
You are a video understanding assistant specialized in analyzing panoramic ERP-format videos.  
Given one frame of a panoramic video, your tasks are:  

1. **Filtering**: Identify if the video should be filtered out.  
Output boolean flags for the following conditions (true if the issue exists, false otherwise):  

- non_ERP_format: The video is **not in ERP (Equirectangular Projection)** panoramic format. For example, if the video looks like a flat perspective, fisheye, cube-map, or any projection other than ERP, set this to true.  

- has_subtitle_or_watermark: The video contains **text overlays, subtitles, logos, or watermarks**. Look carefully for visible text at the bottom, center, or corners of the video. If such elements are present and not part of the real scene, set this to true.  

- edge_missing: The top or bottom edges of the ERP panorama are **cut off, blacked out, cropped, or covered by logos/watermarks**, so the full 360 vertical coverage is missing or obstructed. If you cannot clearly see the poles (sky/ground) or if the edges are hidden by overlays, set this to true.  

- has_overlay: The frame contains **artificial overlays**, such as embedded UI elements, pop-up graphics, stickers, video-in-video inserts, menus, or other synthetic elements that are not part of the natural scene. If you see signs of AR/VR interface, streaming UI, or added images, set this to true.  

- low_quality: The video is of **poor visual quality**, such as being blurry, noisy, heavily pixelated, very low resolution, or distorted in a way that prevents recognizing the scene. If the content is hard to interpret due to quality issues, set this to true.  

- unnatural_content: The video contains **cartoons, animations, CGI, synthetic 3D renderings, or game engine graphics** rather than real-world panoramic footage. If the content is not realistic, set this to true.  

2. **POI Categorization**: From the provided list of categories, select **one or more most relevant** labels that best describe the scene.  
   Only use the given categories, do not invent new ones.  

---

**poi_category list (choose only from below):**

Restaurant, Coffee-Shop, Bars-and-Pubs, Residential-area, Hotels-Motels, Vaccation-Rentals, Hospitals-Clinics, Pharmacies, Dentists, School-Universities, Library, Supermarkets, Shopping-Malls, Clothing-Stores, Shoe-Stores, Bookstores, Flowerstore, Furniture-Stores, Electorical-Store, Pet-Store, Toy-Shop, Airports, Train-Stations, Bus-Stops, Gas-Station, Car-Rental-Agencies, Theaters, Concert-Halls, Sports-Stadiums, Parks-and-Recreation-Areas, Museums, Art-Galleries, Zoos-Aquariums, Botanical-Gardens, Landmarks, Cultural-Centers, Post-Offices, Police-Stations, Courthouses, CityHalls, Banks-ATMs, Events-Conferences-halls, Beaches, Hiking-Trails, Campgrounds, Lakes, Mountains, Forest-Mountains, Farms, Street-View, Square, Business-Centers, Tech-Companies, Co-working-Spaces, Gyms-and-Fitness-Centers, Sports-Clubs, Swimming-Pools, Tennis-Courts, Auto-Repair-Shops, Car-Washes, Parking-Lots, Churches, Mosques, Temples, Graveyards.

---

**Output strictly in JSON format** as follows:

```json
{
  "filter": {
    "non_ERP_format": false,
    "has_subtitle_or_watermark": false,
    "edge_missing": false,
    "has_overlay": false,
    "low_quality": false,
    "unnatural_content": false
  },
  "poi_category": ["Mountains"]
}
```
\end{PromptBox}
\caption{
    Prompt used for panoramic video filtering and scene categorization.
}\label{fig:filter}
\end{figure*}

    \newpage
    \begin{figure*}[t]
    \centering

    \begin{subfigure}[b]{0.95\linewidth}
        \centering
        \includegraphics[width=\linewidth]{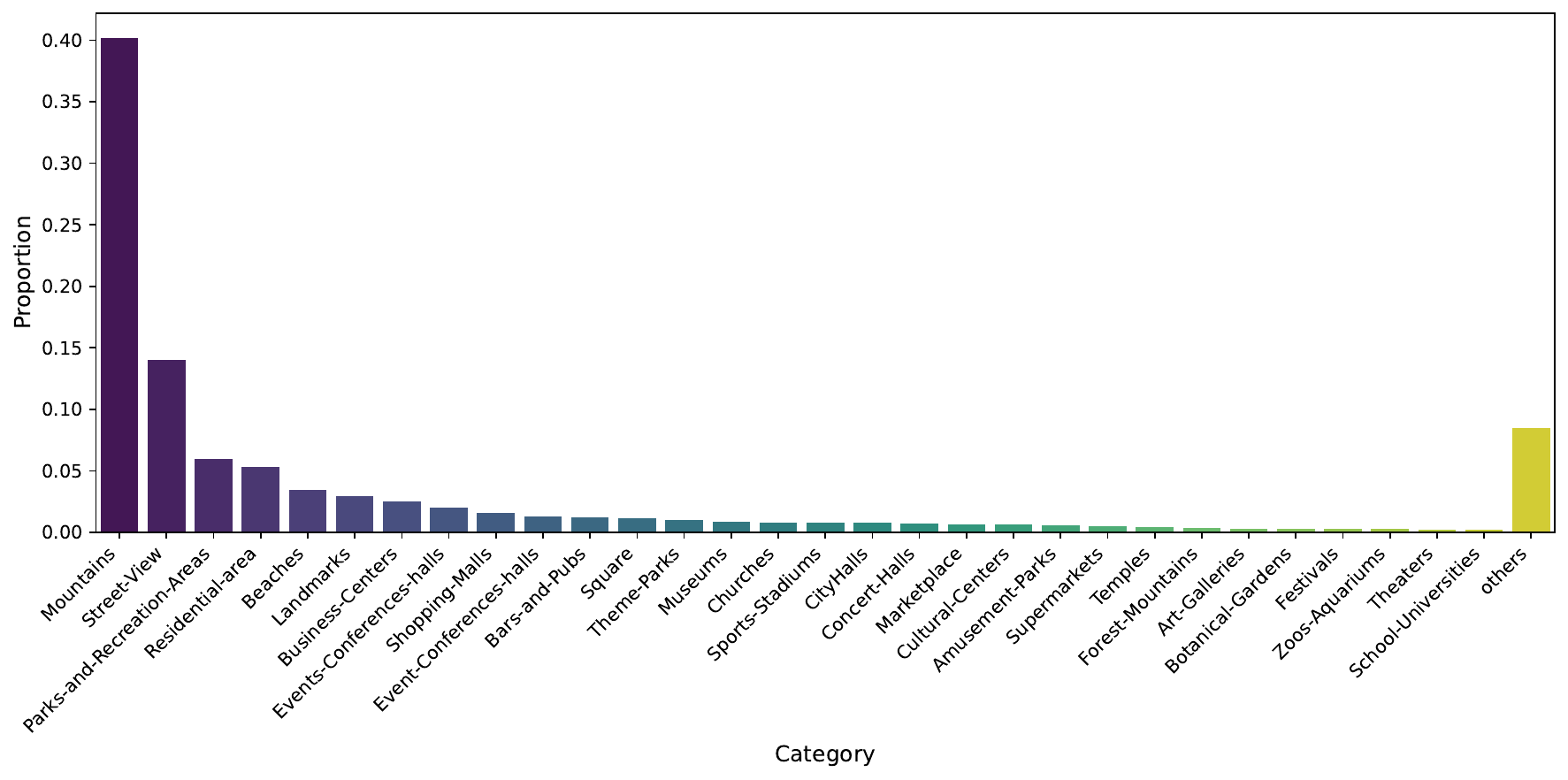}
        \caption{POI category distribution before diversity-constraint trajectory matching.}
        \label{fig:category_before}
    \end{subfigure}

    \vspace{0.7em}

    \begin{subfigure}[b]{0.95\linewidth}
        \centering
        \includegraphics[width=\linewidth]{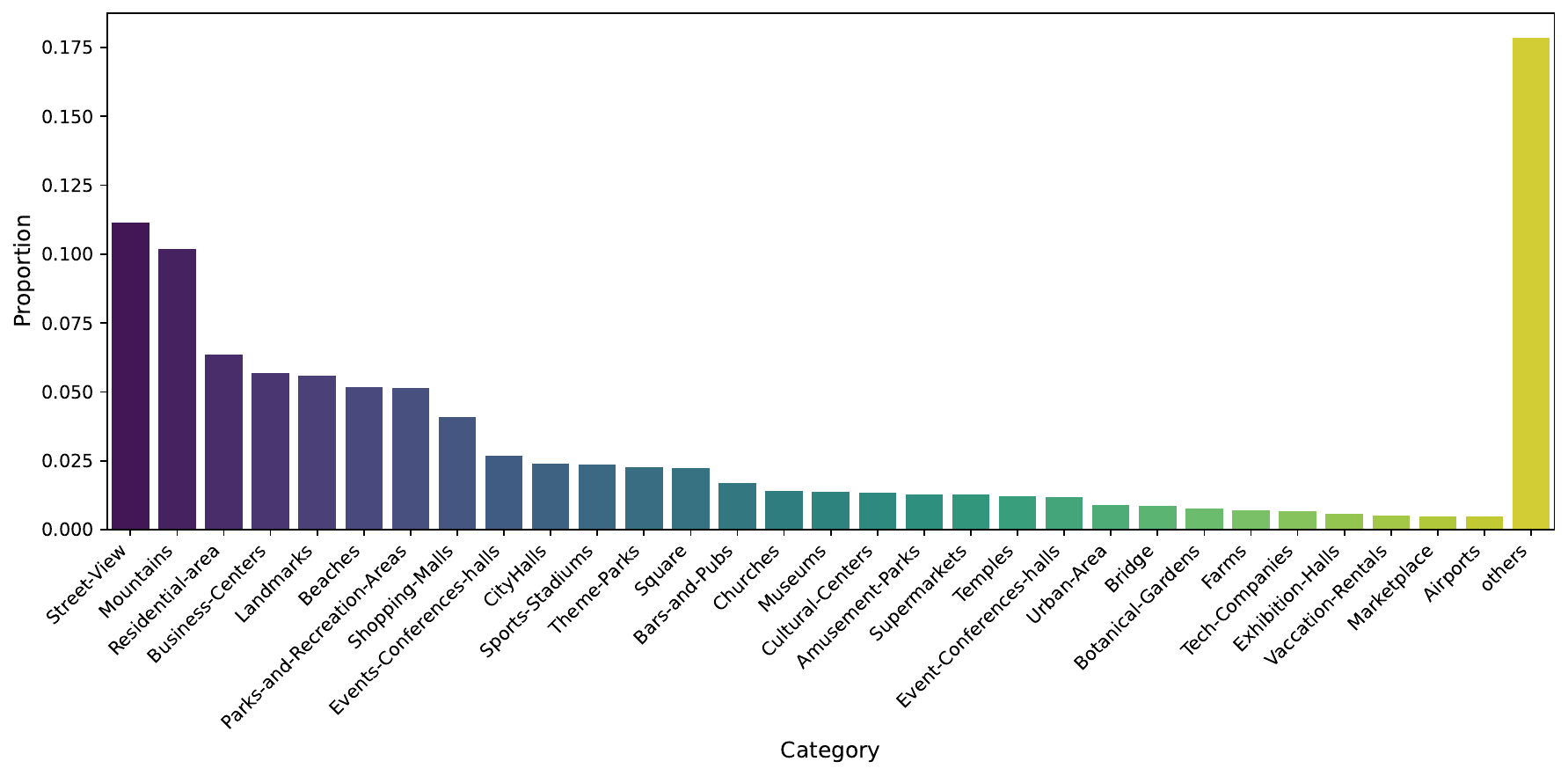}
        \caption{POI category distribution after diversity-constraint trajectory matching.}
        \label{fig:category_after}
    \end{subfigure}

    \caption{
        \textbf{Effect of diversity constraints on POI category distribution.}
        The original 360° panoramic clips exhibit a highly long-tailed distribution across POI categories, dominated by scenes such as mountains and street views (a).
        After applying our diversity constraints during trajectory matching by limiting clip/trajectory reuse and enforcing semantic balance, the resulting dataset becomes more balanced (b), reducing over-represented categories and improving semantic coverage for downstream video synthesis.
    }
    \label{fig:two_level_full_width}
\end{figure*}

    \clearpage
}

\subsection{Video Synthesis with Diverse Cameras}
\label{sec:syn_video}

From the previous stage, each panoramic clip is paired with five transferred rotation trajectories.  
To further enrich motion variation, we apply additional perturbations and panning motions to these trajectories.
We then use the Unified Camera Model (UCM)~\cite{mei2007single} to synthesize camera-diverse videos by projecting the panoramic frames into virtual cameras with varied intrinsics and distortions.

\paragraph{Camera Rotation Augmentation.}
For each transferred trajectory, we generate three augmented variants:
\begin{itemize}
    \item \textbf{Consistent yaw perturbation:} A random yaw offset sampled from \([-180^\circ,180^\circ]\) is added to all frames.
    \item \textbf{Consistent yaw/pitch perturbation:} Random yaw and pitch offsets sampled from \([-180^\circ,180^\circ]\) and \([-80^\circ,80^\circ]\), respectively, are applied uniformly across the clip.
    \item \textbf{Smooth panning motion:} A smooth yaw-pitch-roll panning curve with random start and end offsets is added over the clip duration, using sampling ranges of \([-90^\circ,90^\circ]\) for yaw, \([-40^\circ,40^\circ]\) for pitch, and \([-30^\circ,30^\circ]\) for roll.
\end{itemize}
These augmentations produce three additional rotation variants per trajectory, resulting in a total of \(48\mathrm{k}\) unique camera motions.
Each augmented rotation sequence defines a virtual camera rotation \(\mathbf{R}^{\mathrm{aug}}\in\mathrm{SO}(3)\) for each panoramic frame, used later to project into the desired view.

\paragraph{Unified Camera Model (UCM).}
To synthesize videos with diverse intrinsics and distortions, we adopt the Unified Camera Model (UCM)~\cite{mei2007single}, which represents a wide range of central cameras using a single distortion parameter~\(\xi\).

\textbf{Projection.}
Given a 3D point
\(\bm{p} = [p_x,\, p_y,\, p_z]^\top\)
in the camera coordinate system, UCM first computes:
\begin{equation}
r = \sqrt{p_x^{2} + p_y^{2} + p_z^{2}},
\qquad
\beta = p_z + \xi r,
\label{eq:ucm_r_gamma}
\end{equation}
and then projects the point to pixel coordinates:
\begin{equation}
u = f_x\,\frac{p_x}{\beta} + c_x,
\qquad
v = f_y\,\frac{p_y}{\beta} + c_y,
\label{eq:ucm_projection}
\end{equation}
where \(f_x, f_y, c_x, c_y\) denote the camera intrinsics.
The distortion parameter~\(\xi\) controls the nonlinearity of the projection:
when \(\xi = 0\), UCM reduces to the standard pinhole model,
while larger values of~\(\xi\) introduce stronger bending of rays,
allowing UCM to approximate wide-angle and fisheye lenses within a unified formulation.

\textbf{Ray mapping.}
For a pixel coordinate \((u,v)\), we first obtain normalized image coordinates:
\begin{equation}
x = \frac{u - c_x}{f_x},
\qquad
y = \frac{v - c_y}{f_y},
\label{eq:ucm_normcoord}
\end{equation}
and then lift \((x,y)\) back to a 3D ray direction under UCM:
\begin{equation}
\begin{aligned}
\bm{d}^{\mathrm{cam}}
&=
\frac{1}{\sqrt{x^{2} + y^{2} + \big(1 - \xi \rho\big)^{2}}}
\begin{bmatrix}
x \\[2pt] y \\[2pt] 1 - \xi \rho
\end{bmatrix},\\[6pt]
\rho
&= \sqrt{x^{2} + y^{2}}.
\end{aligned}
\label{eq:ucm_unprojection}
\end{equation}
Since UCM is central, each ray originates from the same center \(\bm{o}^{\mathrm{cam}} = \bm{0}\).

\textbf{FoV-based intrinsic re-parameterization.}
Instead of specifying intrinsics through \(f_x\) and \(f_y\), which is less intuitive, we re-parameterize the camera using its horizontal field of view \(\mathrm{xFoV}\).
For an image width \(W\), the corresponding focal length is:
\begin{equation}
f_x = f_y
=
\frac{W}{2}\,
\frac{\cos\gamma + \xi}{\sin\gamma},
\qquad
\gamma = \tfrac{1}{2}\,\mathrm{xFoV}.
\label{eq:ucm_fov_to_focal}
\end{equation}
Throughout our formulation, we assume the principal point is centered in the image,
\(
c_x = \tfrac{1}{2}W,\;
c_y = \tfrac{1}{2}H,
\)
where \(H\) and \(W\) denote the image height and width.

\textbf{Final ray mapping and projection.}
Using the focal length derived from~\cref{eq:ucm_fov_to_focal}, the UCM unprojection in~\cref{eq:ucm_unprojection} defines, for each pixel \((u,v)\), a central-camera ray
\begin{equation}
\bm{d}^{\mathrm{cam}}_{u,v}
=
\Phi^{\mathrm{UCM}}\!\big(u,\,v;\, \mathrm{xFoV},\, \xi,\, H,\, W\big),
\label{eq:ucm_final_ray}
\end{equation}
where \(\Phi^{\mathrm{UCM}}\) denotes the FoV-parameterized UCM ray-mapping function.

Analogously, the UCM projection of a 3D point
\(\bm{p}\in\mathbb{R}^3\)
to the image plane is given by another mapping
\begin{equation}
(u,v)
=
\Pi^{\mathrm{UCM}}\!\big(\bm{p};\, \mathrm{xFoV},\, \xi,\, H,\, W\big),
\label{eq:ucm_projection_operator}
\end{equation}
which applies~\cref{eq:ucm_projection} using the same FoV-parameterized intrinsics.
These two operators form the basis of UCM and are used throughout the pipeline to generate camera-diverse views.

\paragraph{Camera Intrinsics and Distortion Sampling.}
To cover a broad range of camera geometries, we sample the horizontal field of view \(\mathrm{xFoV}\) and UCM distortion parameter \(\xi\) from several lens-dependent uniform ranges.
We organize cameras into four categories, from pinhole to extreme fisheye, and draw
\(\mathrm{xFoV}\) and \(\xi\) uniformly within their respective intervals:
\begin{itemize}
    \item \textbf{Pinhole:}
    \(\mathrm{xFoV}\!\in[90^\circ,\,110^\circ],\;
    \xi\!\in[0.0,\,0.0]\).
    \item \textbf{Wide-angle:}
    \(\mathrm{xFoV}\!\in[110^\circ,\,140^\circ],\;
    \xi\!\in[0.5,\,0.95]\).
    \item \textbf{Fisheye:}
    \(\mathrm{xFoV}\!\in[140^\circ,\,180^\circ],\;
    \xi\!\in[1.05,\,2.0]\).
    \item \textbf{Extreme fisheye:}
    \(\mathrm{xFoV}\!\in[160^\circ,\,200^\circ],\;
    \xi\!\in[1.5,\,2.3]\).
\end{itemize}
This sampling scheme provides broad coverage of real-world intrinsics and distortion levels, enabling UCPE to learn consistent ray representations across diverse camera types.

\paragraph{Panoramic-to-UCM Projection.}
Given an equirectangular panorama \(I^{\mathrm{ERP}}\in\mathbb{R}^{H' \times W' \times 3}\),
our goal is to project it to a UCM virtual camera view by mapping each UCM ray to its corresponding location on the panorama.

For each pixel \((u,v)\in\{1,\ldots,W\}\times\{1,\ldots,H\}\), we first obtain the FoV-parameterized UCM ray
\(\bm{d}^{\mathrm{cam}}_{u,v}\) using~\cref{eq:ucm_final_ray}.
Applying the virtual camera rotation \(\mathbf{R}^{\mathrm{aug}}\in\mathrm{SO}(3)\) yields the ray expressed in the equirectangular camera frame:
\begin{equation}
\bm{d}^{\mathrm{ERP}}_{u,v}
=
\mathbf{R}^{\mathrm{aug}}\,
\bm{d}^{\mathrm{cam}}_{u,v}.
\label{eq:ucm_ray_world_uv_revised}
\end{equation}
Denoting ray as \(\bm{d}^{\mathrm{ERP}}_{u,v} = [d'_x,\, d'_y,\, d'_z]^\top\), its corresponding location on the panorama is obtained using spherical projection:
\begin{equation}
(u',\, v')
=
\left(
\frac{\operatorname{atan2}(d'_x,d'_z)+\pi}{2\pi},\;
\frac{\arcsin(d'_y)+\tfrac{\pi}{2}}{\pi}
\right).
\label{eq:erp_project_simplified_revised}
\end{equation}

The final UCM-rendered frame is obtained by sampling the panorama at these coordinates:
\begin{equation}
I^{\mathrm{UCM}}[u,v]
=
I^{\mathrm{ERP}}\!\big(u',\, v'\big),
\label{eq:erp_sample_simplified_revised}
\end{equation}
resulting in a UCM image consistent with the sampled intrinsics \((\mathrm{xFoV},\xi)\) and the virtual camera pose.
This process is repeated for all frames in the panoramic clip to produce the final synthesized video.

\paragraph{Virtual Camera Pose Composition.}
For each rendered frame, the equirectangular SLAM system provides a camera-to-world pose
\(\mathbf{T}^{\mathrm{ERP}}\in\mathrm{SE}(3)\),
which we compose with the augmented rotation
\(\mathbf{R}^{\mathrm{aug}}\in\mathrm{SO}(3)\)
to obtain the final virtual-camera pose.
Let
\(\mathbf{R}^{\mathrm{ERP}}\)
and
\(\mathbf{t}^{\mathrm{ERP}}\)
denote the rotation and translation of
\(\mathbf{T}^{\mathrm{ERP}}\).
The virtual UCM pose is then
\begin{equation}
\mathbf{T}^{\mathrm{UCM}}
=
\begin{bmatrix}
\mathbf{R}^{\mathrm{ERP}}\,\mathbf{R}^{\mathrm{aug}}
&
\mathbf{t}^{\mathrm{ERP}}
\\[2pt]
\mathbf{0}^\top & 1
\end{bmatrix},
\label{eq:ucm_pose_final}
\end{equation}
which replaces only the orientation while preserving the original trajectory
translation.
This pose is paired with the sampled intrinsics \((\mathrm{xFoV},\xi)\) for camera-aware training.

\paragraph{UCM Video Captioning.}
To provide descriptive captions for each synthesized video for text-to-video training, we adapt vLLM model~\cite{bai2025qwen2}
with a simple prompt:
\begin{PromptBox}
You are a helpful video captioning assistant.
Please describe this video in detail.
\end{PromptBox}

\paragraph{}
In total, we generate approximately \(48\mathrm{k}\) 81-frames video clips at a resolution of \(480 \times 832\) and \(16~\mathrm{fps}\). The entire generation process takes approximately 3 days on a single A800 GPU node.
For evaluation, we match the official test split of CameraBench with the rest of 360° videos and enforce stricter diversity constraints, resulting in \(272\) test clips.
We note that while our dataset is synthesized with UCM, our UCPE representation is compatible with other camera models, as it encodes rays in a model-agnostic manner.

\subsection{Camera Motion and Intrinsics Statistics}

\begin{figure}[t]
    \centering

    \begin{subfigure}[b]{0.48\linewidth}
        \centering
        \includegraphics[width=\linewidth]{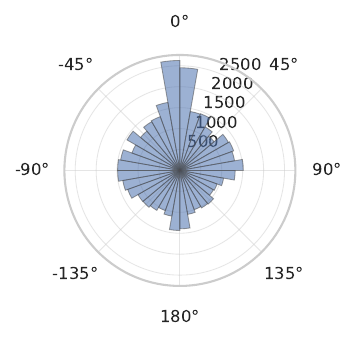}
        \caption{Our dataset}
        \label{fig:direction_panshot}
    \end{subfigure}
    \hfill
    \begin{subfigure}[b]{0.48\linewidth}
        \centering
        \includegraphics[width=\linewidth]{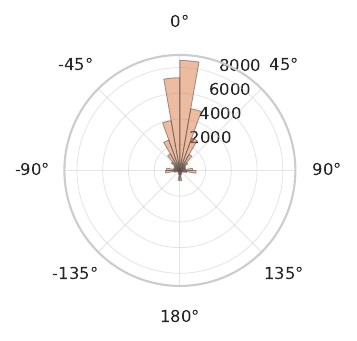}
        \caption{RealEstate10K}
        \label{fig:direction_re10k}
    \end{subfigure}

    \caption{
        \textbf{Camera Motion Direction Distribution.}
        We visualize the horizontal translation directions using rose plots for (a) our dataset and (b) RealEstate10K.
        Compared with the strong forward-motion bias in RealEstate10K, our dataset provides substantially richer and more uniformly distributed camera translation directions.
    }
    \label{fig:direction}
\end{figure}

\begin{figure}[t]
    \centering

    \begin{subfigure}[b]{0.48\linewidth}
        \centering
        \includegraphics[width=\linewidth]{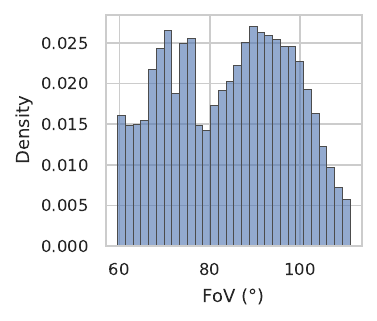}
        \caption{Our dataset}
        \label{fig:fov_panshot}
    \end{subfigure}
    \hfill
    \begin{subfigure}[b]{0.48\linewidth}
        \centering
        \includegraphics[width=\linewidth]{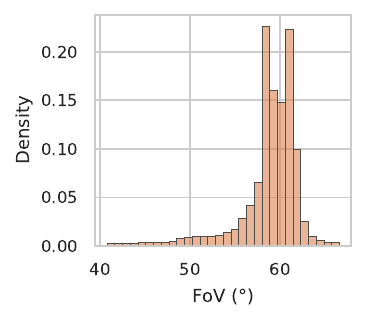}
        \caption{RealEstate10K}
        \label{fig:fov_re10k}
    \end{subfigure}

    \caption{
        \textbf{Camera Field of View Distribution.}
        We compare the vertical FoV distributions of (a) our dataset and (b) RealEstate10K.
        Our dataset exhibits a wide FoV range (60°-110°), reflecting the diverse intrinsics sampled during synthesis.
        In contrast, RealEstate10K shows a narrow distribution concentrated around 60°, indicating limited variability in camera intrinsics.
    }
    \label{fig:fov}
\end{figure}

\paragraph{Camera Motion Direction Distribution.}
To assess the diversity of camera motion in our dataset, we visualize the distribution of translation directions between the first and last frames, as shown in~\cref{fig:direction}.
We compare our synthetic dataset against RealEstate10K~\cite{zhou2018stereo}, a real-world collection containing approximately \(66\mathrm{k}\) clips in the training split.
For fair comparison, we discard clips with fewer than \(81\) frames and use the first \(81\) frames of the remaining videos, yielding \(47\mathrm{k}\) clips.
As shown in~\cref{fig:direction_panshot} and~\cref{fig:direction_re10k}, our dataset covers a more balanced range of translation directions, largely due to the consistent yaw perturbation introduced during augmentation, whereas RealEstate10K is dominated by forward translations characteristic of real-estate tours.

\paragraph{Field of View Distribution.}
We further compare the vertical field-of-view (FoV) distributions of our dataset and RealEstate10K.
As shown in~\cref{fig:fov}, our dataset spans a broad FoV range from 60° to 110°, owing to the diverse intrinsics sampled during synthesis.
In contrast, RealEstate10K is largely concentrated near 60°, reflecting its limited intrinsic variability.
Additionally, our dataset incorporates a wide spectrum of distortion levels under the Unified Camera Model (UCM), covering multiple lens types, whereas RealEstate10K primarily contains distortion-free pinhole cameras.

\section{Details of Absolute Orientation Encoding}
\label{sec:abs_orient}

As introduced in Sec.~3.2 of the main paper, the Lat-Up representation consists
of a latitude map and an Up map. The latitude component is computed directly
from world-space ray directions using Eq.~8 of the main paper. This section
details the derivation of the Up map.

For each token \(t\), let \((u_t, v_t)\) denote its pixel center in the image
plane. Using the FoV-parameterized UCM intrinsics collected in \(\phi\), the
camera-frame ray direction is obtained via the UCM ray-mapping operator
defined in \cref{eq:ucm_final_ray}:
\begin{equation}
\bm{d}^{\mathrm{cam}}_t
=
\Phi^{\mathrm{UCM}}_{\phi}\!\big(u_t, v_t\big),
\qquad
\|\bm{d}^{\mathrm{cam}}_t\| = 1.
\end{equation}
Applying the camera-to-world rotation \(\mathbf{R}\) yields the world-frame ray
direction (Eq.~1 in the main paper):
\begin{equation}
\bm{d}_t = \mathbf{R}\,\bm{d}^{\mathrm{cam}}_t,
\qquad
\|\bm{d}_t\| = 1.
\end{equation}

\paragraph{Ray perturbation direction towards world up.}
The Up map is constructed by examining how each viewing ray \(\bm{d}_t\)
responds to an infinitesimal perturbation toward the world up axis
\(\bm{u}^{\mathrm{wld}}=[0,-1,0]^\top\).
Intuitively, this perturbation reflects how the ray would move on the unit
viewing sphere if it were slightly rotated in the direction of world up, and
its image-plane displacement reveals the 2D Up direction associated with token
\(t\).

On the sphere, the instantaneous motion of the ray under such a perturbation
must lie in the tangent direction orthogonal to both \(\bm{d}_t\) and
\(\bm{u}^{\mathrm{wld}}\).  
This tangent direction is given by the cross product:
\begin{equation}
\bm{k}_t = \bm{d}_t \times \bm{u}^{\mathrm{wld}},
\qquad
\hat{\bm{k}}_t = \frac{\bm{k}_t}{\|\bm{k}_t\|}.
\label{eq:up_axis_suppl}
\end{equation}
The unit vector \(\hat{\bm{k}}_t\) thus specifies the correct axis for applying
an infinitesimal rotation to \(\bm{d}_t\) toward world up, forming the basis for
the Up map defined in the following steps.

\paragraph{Small-angle perturbation via Rodrigues' formula.}
We rotate \(\bm{d}_t\) by a small fixed angle \(\delta\) (e.g., \(\delta=0.1\) rad)
around \(\hat{\bm{k}}_t\).  
Using Rodrigues' formula, the perturbed world-space ray direction is
\begin{align}
\bm{d}^{\mathrm{rot}}_t
&=
\operatorname{Rot}\!\big(\hat{\bm{k}}_t,\delta\big)\,\bm{d}_t
\\[3pt]
&=
\bm{d}_t \cos\delta
+
\big(\hat{\bm{k}}_t \times \bm{d}_t\big)\sin\delta
+
\hat{\bm{k}}_t\,(\hat{\bm{k}}_t^\top \bm{d}_t)\,(1-\cos\delta).
\label{eq:up_rodrigues}
\end{align}

We then transform this perturbed ray back to the camera frame:
\begin{equation}
\bm{d}^{\mathrm{cam,rot}}_t = \mathbf{R}^\top \bm{d}^{\mathrm{rot}}_t,
\end{equation}
and project it with the UCM projection operator
\(\Pi^{\mathrm{UCM}}_{\phi}\) introduced in \cref{eq:ucm_projection_operator}:
\begin{equation}
\big(u^{\mathrm{rot}}_t, v^{\mathrm{rot}}_t\big)
=
\Pi^{\mathrm{UCM}}_{\phi}\!\big(\bm{d}^{\mathrm{cam,rot}}_t\big).
\end{equation}

Importantly, the original \((u_t, v_t)\) is the pixel location of token \(t\),
while \((u^{\mathrm{rot}}_t, v^{\mathrm{rot}}_t)\) is the projected position of
the ray after a small rotation toward the world up direction.

\paragraph{Up map definition.}
The induced image-plane displacement is
\begin{equation}
\Delta u_t = u^{\mathrm{rot}}_t - u_t,
\qquad
\Delta v_t = v^{\mathrm{rot}}_t - v_t,
\end{equation}
and the Up map at token \(t\) is defined as
\begin{equation}
\mathrm{Up}_t
=
\frac{
\left[\,\Delta u_t,\;\Delta v_t\,\right]
}{
\sqrt{\Delta u_t^2 + \Delta v_t^2}
}.
\label{eq:up_final_suppl}
\end{equation}

Beyond its geometric definition, the resulting Up map provides a strong appearance-dependent cue: its spatial pattern varies coherently with the camera's absolute pitch and roll, and also responds to scene semantics such as the vertical structure of buildings and trees.
Moreover, because the perturbation is evaluated through the projection fuction \(\Pi\), the Up map naturally reflects the characteristic warping of different lenses.
As a result, it offers a unified visual signal that jointly encodes global orientation, semantic regularities, and lens-induced distortions.

\section{Evaluation Metrics}
\label{sec:eval}

In Sec.~4.1 of the main paper, we evaluate both video generation quality and
camera controllability.  
Here, we detail the computation of camera-related metrics used for benchmarking
relative pose control, absolute orientation, and lens controllability.

\paragraph{Relative Camera Pose Control.}

For each generated video, we begin by rectifying every frame to a pinhole projection.  
This step is necessary because existing pose estimation methods, including ViPE~\cite{huang2025vipe}, are not robust under strong lens distortions and wide-FoV warping, which frequently appear in our synthesized UCM views.
Using the ground-truth UCM parameters \((\mathrm{xFoV}, \xi)\), each distorted frame is mapped to a pinhole image whose effective horizontal FoV is capped at \(100^\circ\).
Rectification is implemented using the UCM ray-pixel operators \(\Phi^{\mathrm{UCM}}_{\phi}\) and \(\Pi^{\mathrm{UCM}}_{\phi}\), which providea forward-inverse mapping between distorted pixels and their underlying 3D
rays.
The rectified sequence is then fed to ViPE to estimate camera-to-world trajectories \(\{\hat{\mathbf{T}}^{\mathrm{wc}}_i\}\).

Following~\cite{he2024cameractrl, wang2024motionctrl}, all pose metrics operate on \emph{relative} trajectories.  
For a camera-to-world pose
\(
\mathbf{T}^{\mathrm{wc}}_i
=
\left[
\begin{smallmatrix}
\mathbf{R}^{\mathrm{wc}}_i & \mathbf{t}^{\mathrm{wc}}_i \\
\mathbf{0}^\top & 1
\end{smallmatrix}
\right]
\in SE(3),
\)
with rotation \(\mathbf{R}_i\) and translation \(\mathbf{t}_i\), the corresponding relative ground truth pose and estimated pose is defined as
\begin{equation}
\mathbf{T}_i
=
(\mathbf{T}^{\mathrm{wc}}_0)^{-1}
\mathbf{T}^{\mathrm{wc}}_i,
\qquad
\hat{\mathbf{T}}_i
=
(\hat{\mathbf{T}}^{\mathrm{wc}}_0)^{-1}
\hat{\mathbf{T}}^{\mathrm{wc}}_i.
\end{equation}

\begin{itemize}

    \item \textbf{Rotation Error (RotErr):}
    The total rotation error is computed as the sum of per-frame angular
    deviations:
    \begin{equation}
    \mathrm{RotErr}
    =
    \sum_i
    \arccos\!\left(
        \frac{
            \mathrm{Tr}\!\left(
                \mathbf{R}_i^{\top}\hat{\mathbf{R}}_i
            \right)-1
        }{2}
    \right).
    \end{equation}

    \item \textbf{Translation Error (TransErr):}
    For relative translations \(\mathbf{t}_i\) and \(\hat{\mathbf{t}}_i\),
    \begin{equation}
    \mathrm{TransErr}
    =
    \sum_i
    \big\|
        \mathbf{t}_i - \hat{\mathbf{t}}_i
    \big\|_2 .
    \end{equation}

    \item \textbf{Camera Motion Consistency (CamMC):}
    Flatten first three rows of each relative pose into a 12-dim vector and compute
    \begin{equation}
    \mathrm{CamMC}
    =
    \sum_i
    \big\|
        \operatorname{Vec}(\mathbf{T}_i)
        -
        \operatorname{Vec}(\hat{\mathbf{T}}_i)
    \big\|_2 .
    \end{equation}

\end{itemize}

These metrics evaluate how closely the generated trajectory follows the target camera motion, independent of absolute scale or global alignment.

Since different baselines generate videos with varying lengths and frame rates (\eg, CameraCtrl produces 16 frames at 4\,fps, AC3D produces 48 frames at 8\,fps, while our method outputs 81 frames at 16\,fps), a direct trajectory comparison would be biased by temporal sampling.  
To ensure a fair evaluation of relative camera pose control, we uniformly sample the same 16 timestamps across all methods to compute the pose metrics.

\paragraph{Absolute Camera Orientation and Lens Control.}

We employ GeoCalib~\cite{veicht2024geocalib} to estimate per-frame absolute orientation and lens parameters from generated frames.
For each frame \(i\), GeoCalib outputs the predicted pitch \(\hat{\theta}_i\) and roll \(\hat{\varphi}_i\), the vertical field of view \(\widehat{\mathrm{FoV}}\), and the radial distortion coefficients \(\hat{k}_1, \hat{k}_2\) under the classical radial model~\cite{Brown1966DecenteringDO}.  

Ground-truth parameters \(({\theta}_i^{\ast}, {\varphi}_i^{\ast}, \mathrm{FoV}^{\ast}, k_1^{\ast}, k_2^{\ast})\) are obtained from the GeoCalib estimation on the original UCM frames.
We then compute the following metrics:

\begin{itemize}

    \item \textbf{Pitch / Roll Error:}  
    We compute the absolute error
    \begin{equation}
    |\hat{\theta}_i - \theta_i^{\ast}|,
    \qquad
    |\hat{\varphi}_i - \varphi_i^{\ast}|.
    \end{equation}

    \item \textbf{FoV and Distortion Errors:}
    Under the radial Brown model~\cite{Brown1966DecenteringDO}, we compute
    \begin{equation}
    |\hat{\mathrm{FoV}} - \mathrm{FoV}^\ast|,
    \qquad
    |\hat{k}_1 - k_1^\ast|,
    \qquad
    |\hat{k}_2 - k_2^\ast|.
    \end{equation}

\end{itemize}

Together, these metrics measure controllability over absolute camera orientation, lens intrinsics, and distortion.
providing a detailed evaluation of camera-aware generation quality.

\section{Implementation Details}
\label{sec:impl}

\begin{figure*}[t]
    \centering
    \includegraphics[width=1.\linewidth, trim={0 0 0 0}, clip]{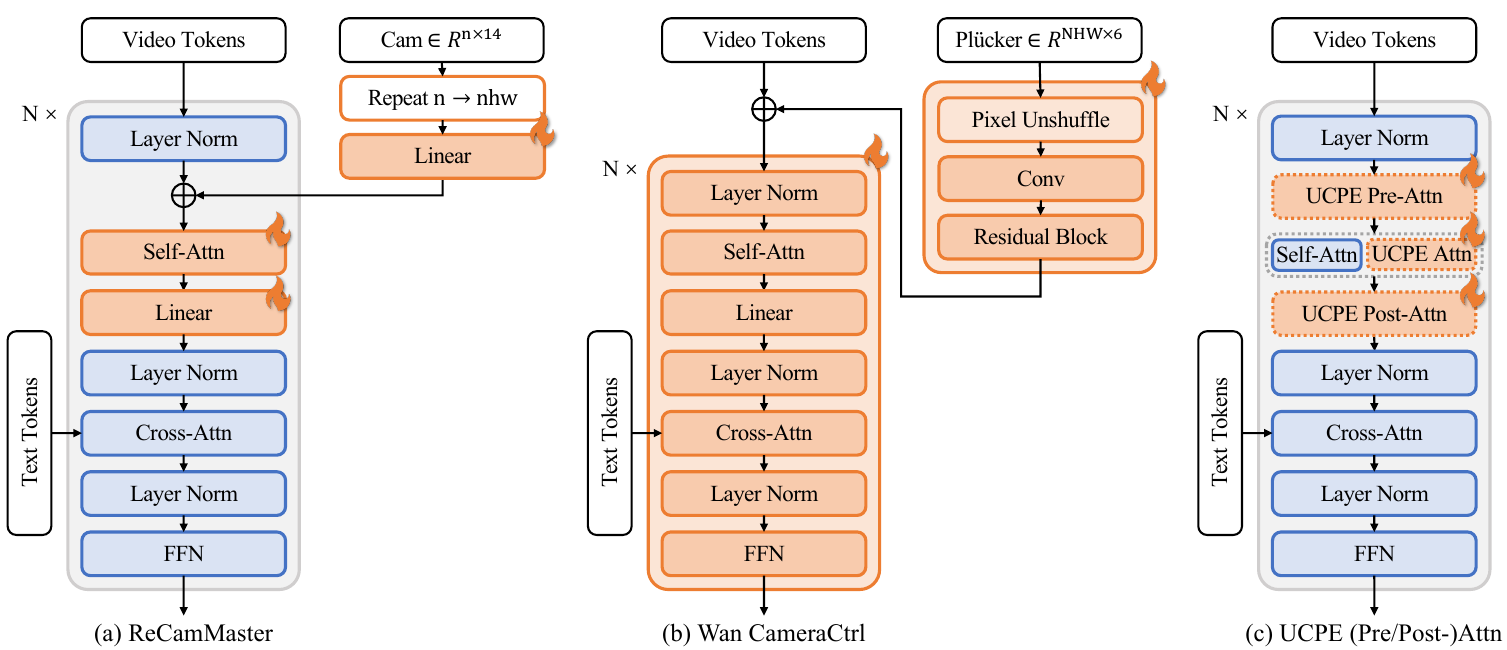}
    \caption{
        \textbf{Implementation of Baseline And Ablation Models.}
        (a) ReCamMaster injects per-frame camera parameters into each Transformer block after spatial repetition. 
        (b) Wan CameraCtrl injects Pl\"ucker-encoded rays into video tokens using a convolutional adapter.
        (c) Our UCPE ablations insert a spatial attention adapter before, after, or in parallel with the original self-attention module.
    }\label{fig:baselines}
\end{figure*}

\paragraph{Our Method.}
We adopt Wan2.1-T2V-1.3B~\cite{wan2025wan} as our base model.
It consists of a text encoder, a Diffusion Transformer, and a 3D VAE, totaling \(7.3\) billion parameters.
We fine-tune the model on our synthesized dataset while freezing all original weights, and train only the proposed attention adapters using AdamW~\cite{loshchilovdecoupled} with a learning rate of \(1\mathrm{e}{-4}\) for 10k steps.
Training uses a batch size of 8 on 8 NVIDIA A800 GPUs and completes within roughly one day.
At inference time, we generate \(81\) frames at a resolution of \(480 \times 832\) and \(16~\mathrm{fps}\).
Since no existing text-to-video model supports full camera control (\ie, relative pose, absolute orientation, and lens parameters), we adapt two recent methods for comparison, as detailed below.

\paragraph{ReCamMaster.}
We adapt the official ReCamMaster implementation~\cite{bai2025recammaster} for text-to-video generation with complete camera control.
As illustrated in~\cref{fig:baselines}a, ReCamMaster injects raw camera parameters into every Transformer block.
In our adaptation, we concatenate the normalized field of view \(\mathrm{xFoV}'=\mathrm{xFoV}/180^\circ\) and the distortion parameter \(\xi\) with the flattened \(12\)-dim pose vector \(\operatorname{Vec}(\mathbf{T}^{\mathrm{wc}}_i)\) for each frame \(i\).
Because Wan2.1-T2V-1.3B applies a \(4\times\) temporal compression in the VAE, these per-frame parameters are downsampled to match the token sequence length \(n\).
Following the original design, the camera parameters are then broadcast across all spatial tokens (\(h \times w\)), yielding a tensor of shape \((nhw, 14)\), projected with zero-initialized linear layers to the token dimension, and added as a bias before each self-attention layer.
A linear layer with identity initialization performs feature projection after self-attention.
During training, we fine-tune the added linear layers and the original self-attention layers following the ReCamMaster training scheme.

\paragraph{Wan CameraCtrl.}
We further adapt Wan2.1-Fun-V1.1-1.3B-Control-Camera, a third-party implementation of CameraCtrl~\cite{he2024cameractrl} built on Wan~\cite{wan2025wan}, for text-to-video generation.
As shown in~\cref{fig:baselines}b, Wan CameraCtrl converts Pl\"ucker-encoded rays of shape \((NHW, 6)\), where \(N\) is the number of frames and \(H, W\) denote spatial resolution of input video, into token-shaped features using a convolutional adapter with pixel-unshuffle downsampling, and injects them via additive biasing.
We retain this structure but generalize the Pl\"ucker encoding from pinhole cameras to UCM cameras using the ray formulation in~\cref{eq:ucm_final_ray}.
For stable training, the convolution and final residual layers are initialized to zero.
We fine-tune all parameters of the convolutional adapter and the diffusion transformer using a learning rate of \(1\mathrm{e}{-5}\), following the official training recipe.

\paragraph{UCPE Spatial Adapters Ablation.}
As described in Sec.~3.3 of the main paper, we insert a spatial attention adapter as a parallel branch to each Diffusion Transformer self-attention layer (see Fig.~3 of the main paper).
In Sec.~4.3, we additionally evaluate two variants that place the adapter immediately before or after the original self-attention.
The three variants are summarized in~\cref{fig:baselines}c and share the same \(1/8\) compression ratio.
All models are trained under the same setting described above.

\paragraph{Inference Latency.}
We measure the inference latency of UCPE and other baselines on an NVIDIA A800 GPU, as summarized in \cref{tbl:latency}. 
UCPE introduces minimal computational overhead compared to the base Wan2.1 model and other camera-conditioned adapters, while generating high-resolution videos with a significantly larger number of frames than earlier methods such as CameraCtrl and AC3D. 
Specifically, UCPE generates 81 frames at $480 \times 832$ resolution in 184 seconds, which is comparable to ReCamMaster (179s) and Wan CameraCtrl (177s) under the same settings.

\begin{table}[t]
    \centering
    \caption{
        \textbf{Inference latency comparison.}
        Latencies are measured on a single NVIDIA A800 GPU, with corresponding frame counts and resolutions provided for reference.
    }\label{tbl:latency}
    \vspace{-0.5em}
    \resizebox{0.75\columnwidth}{!}{
        \begin{tabular}{l|c|c|c}
            \toprule
            Method & Time (s) & Frames & Resolution \\
            \midrule
            ReCamMaster & 179 & 81 & \(480 \times 832\) \\
            Wan CameraCtrl & 177 & 81 & \(480 \times 832\) \\
            CameraCtrl & 17 & 16 & \(256 \times 384\) \\
            AC3D & 632 & 48 & \(480 \times 720\) \\
            UCPE & 184 & 81 & \(480 \times 832\) \\
            \bottomrule
        \end{tabular}
    }
\end{table}

\section{Robustness, Generalization, Limitations, and Future Works}
\label{sec:limitation}

\begin{figure}[t]
    \centering
    \includegraphics[width=1.0\columnwidth]{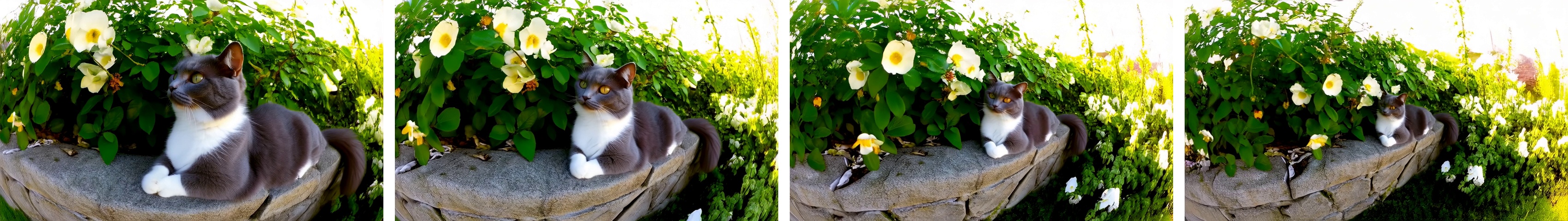}
    \caption{\textbf{Robustness to Text-Camera Conflict.} Even when the text prompt specifies a ``telephoto'' view, UCPE correctly synthesizes a fisheye video according to the provided camera parameters, resolving the conflict between text and geometry.}
    \label{fig:conflict}
\end{figure}

\paragraph{Robustness to Text-Camera Conflict.}
We further evaluate UCPE's robustness when textual prompts conflict with the specified camera geometry.
As shown in \cref{fig:conflict}, we test a prompt specifying a ``flat, telephoto portrait'' of a cat against a contradictory fisheye camera (\(\textrm{xFoV} = 180^\circ, \xi = 2.0\)).
Despite the text suggesting a narrow field of view and minimal distortion, UCPE successfully enforces the requested fisheye geometry, demonstrating that the camera positional encoding provides strong geometric guidance that can override conflicting appearance cues in the text.

\begin{figure}[t]
    \centering
    \includegraphics[width=1.0\columnwidth]{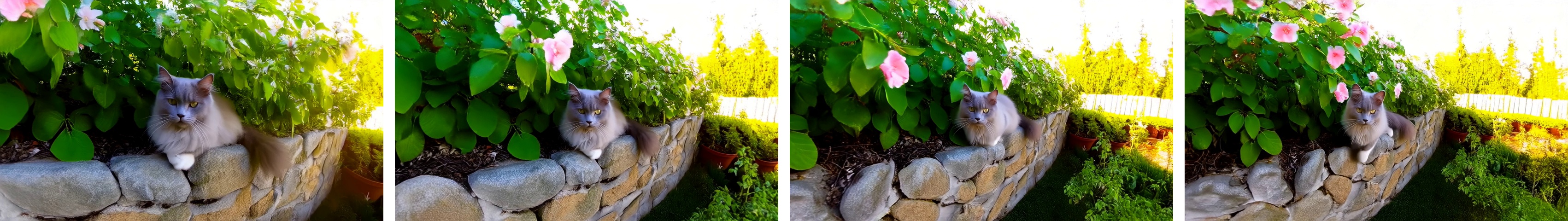}
    \caption{\textbf{Generalization to Unseen Camera Models.} Without fine-tuning, UCPE generalizes to unobserved models such as the Brown-Conrady model by simply substituting the ray-mapping function at inference time.}
    \label{fig:brown_conrady}
\end{figure}

\paragraph{Generalization to Unseen Camera Models.}
UCPE is model-agnostic as it learns universal ray-space interactions.
For unobserved lens types, one can simply replace the ray-mapping function at inference to achieve control without fine-tuning, as the attention mechanism reasons about ray relationships rather than specific camera parameters (\eg, \(\xi\)).
We evaluate this capability on the complex Brown-Conrady camera model used in OpenCV for physical lens calibration (\(\textrm{xFoV} = 100^\circ, k_1 = -0.2, k_2 = -0.05\)).
As shown in \cref{fig:brown_conrady}, UCPE successfully generalizes to this model, adding realistic barrel distortion to the synthesized frames without requiring additional training.

\begin{figure}[t]
    \centering
    \includegraphics[width=1.0\columnwidth]{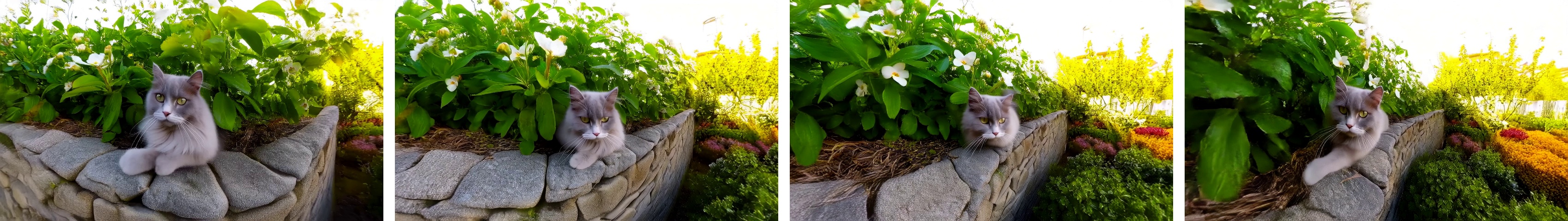}
    \caption{\textbf{Failure Case on Equirectangular Projection.} Testing with ERP ray-mapping results in artifacts as the model was not trained on such extreme panoramic projections.}
    \label{fig:failure_erp}
\end{figure}

\paragraph{Limitations.}
UCPE relies on camera poses during training and currently models only pose, intrinsics, and distortion, without capturing richer attributes such as zoom, focus, or depth-of-field.
Furthermore, we test UCPE with the ray-mapping function of Equirectangular projection (ERP), representing an extreme case not seen during training.
As shown in \cref{fig:failure_erp}, while UCPE fails to synthesize full 360° panoramas correctly under this extreme setting, we expect its performance to improve with targeted fine-tuning on panoramic datasets.

\begin{figure}[t]
    \centering
    \includegraphics[width=1.0\columnwidth]{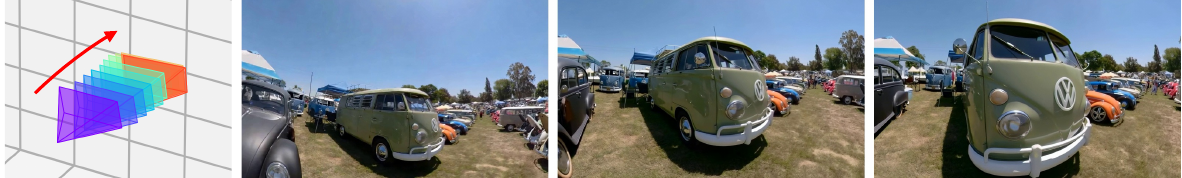}
    \caption{\textbf{Application to I2V Tasks.} We fine-tune Wan2.1-I2V-14B to demonstrate the feasibility of UCPE on more constrained Image-to-Video tasks.}
    \label{fig:supp_i2v}
\end{figure}

\paragraph{Future Works.}
Extending UCPE to support these additional controls and further reducing reliance on accurate pose supervision (\eg, through self-supervised geometric losses) remain promising directions. 
Furthermore, while T2V generation is our primary focus, UCPE is also applicable to Image-to-Video (I2V) tasks. 
In contrast to T2V, I2V tasks benefit from the strong rigid pixel constraints of the input image, which naturally anchors the lens and initial orientation. 
Consequently, UCPE supports the I2V task without the need for Absolute Orientation Encoding (AOE): once the lens is defined (\eg, via GeoCalib), Relative Ray Encoding (RRE) alone can drive subsequent frames. 
To demonstrate this feasibility, we fine-tune Wan2.1-I2V-14B. 
As shown in~\cref{fig:supp_i2v}, our representations readily extend to control the camera pose in more constrained I2V settings. 
Video-to-Video (V2V) tasks (\eg, ReCamMaster) could similarly benefit from UCPE's enhanced controllability. 
Since such tasks inherently require synthetic multi-trajectory datasets, we leave this exploration to future work.

\section{Supplementary Video}
\label{sec:demo}

The supplementary video provides visualizations and demonstrations of the following aspects:
\begin{itemize}
    \item Overview of UCPE (Unified Camera Positional Encoding) and its two components: Relative Ray Encoding and Absolute Orientation Encoding.
    \item Demonstration of controllability over intrinsics and distortions.
    \item Demonstration of controllability over extrinsics and initial camera orientation.
    \item Applications enabled by UCPE, including cinematic content creation, autonomous driving, and embodied AI.
    \item Examples from our large-scale synthetic dataset with diverse intrinsics, distortion profiles, and camera motions (Sec.~4.1 of the main paper, Sec.~A of the supplementary).
    \item Comparison with baseline methods on our synthetic dataset for complete camera-controlled video generation (Fig.~4 of the main paper).
    \item Generalization results on the RealEstate10K dataset, showcasing improved camera controllability and video quality under pinhole camera settings (Fig.~5 of the main paper).
\end{itemize}

\end{document}